\pdfoutput=1

\documentclass[11pt]{article}

\usepackage[]{ACL2023}
\usepackage{hyperref}

\usepackage{times}
\usepackage{latexsym}
\usepackage{CJKutf8}
\usepackage[T1]{fontenc}

\usepackage{booktabs,tabularx}
\usepackage{longtable}
\usepackage{adjustbox}
\usepackage{subcaption}
\usepackage[T1]{fontenc}
\usepackage{multirow}
\usepackage{booktabs}

\usepackage[utf8]{inputenc}

\usepackage{microtype}

\usepackage{inconsolata}

\title{The Next Chapter: A Study of Large Language Models in Storytelling}

\author{Zhuohan Xie 
 \qquad
 Trevor Cohn\thanks{~~Now at Google DeepMind.} 
 \qquad
 Jey Han Lau \\
 School of Computing and Information Systems \\
 The University of Melbourne \\
 zhuohanx@student.unimelb.edu.au, 
\{t.cohn, laujh\}@unimelb.edu.au
}

\begin{document}
\begin{CJK*}{UTF8}{gbsn}
\maketitle
\begin{abstract}
To enhance the quality of generated stories, recent story generation models have been investigating the utilization of higher-level attributes like plots or commonsense knowledge. The application of prompt-based learning with large language models (LLMs), exemplified by GPT-3, has exhibited remarkable performance in diverse natural language processing (NLP) tasks.
This paper conducts a comprehensive investigation, utilizing both automatic and human evaluation, to compare the story generation capacity of LLMs with recent models across three  datasets with variations in style, register, and length of stories. The results demonstrate that LLMs generate stories of significantly higher quality compared to other story generation models. Moreover, they exhibit a level of performance that competes with human authors, albeit with the preliminary observation that they tend to replicate real stories in situations involving world knowledge, resembling a form of  plagiarism.

\end{abstract}

\section{Introduction}

Automatic story generation poses a significant challenge as it requires more than just individually coherent sentences. A good story should exhibit a natural flow, adhere to commonsense logic, and be captivating to the reader.
In recent times, the prevailing approach in story generation involves fine-tuning pre-trained language models (PLMs) like GPT-2 \citep{radford2019language} or BART \citep{lewis-etal-2020-bart} on specific datasets. These models generally excel in generating fluent sentences, devoid of glaring grammar issues. However, they often struggle to construct a coherent story that adheres to commonsense and fails to create an engaging narrative \citep{see-etal-2019-massively, guan-etal-2021-long}. To overcome these challenges, state-of-the-art (SOTA) story generation models integrate higher-level features, such as plots and commonsense knowledge.

Prompt-based learning \citep{DBLP:journals/corr/abs-2107-13586} is a recent paradigm specifically tailored for large language models (LLMs) that possess in-context learning capabilities \citep{DBLP:conf/nips/BrownMRSKDNSSAA20, zhao2023survey}. In contrast to the conventional ``pre-train and fine-tune'' approach, which necessitates a substantial amount of data for fine-tuning, prompt-based learning enables LLMs to learn a task by providing them with multiple examples as a ``prompt'', eliminating the need for gradient-based fine-tuning \citep{DBLP:journals/corr/abs-2107-13586}.
Recently, LLMs have demonstrated remarkable performance across various language generation tasks, with notable attention on models such as ChatGPT and GPT-4 \citep{DBLP:journals/corr/abs-2302-06476, DBLP:journals/corr/abs-2304-01852, DBLP:journals/corr/abs-2303-08774-GPT-4-report}. For instance, a comparative analysis by \citet{DBLP:journals/corr/abs-2302-06476} highlighted the superior performance of fine-tuned LLMs over smaller pre-trained models in zero-shot scenarios for tasks like dialogue and summarization. It is worthwhile to note though, that story generation was not specifically examined in their experiments.


This paper aims to address this research gap by conducting a comprehensive evaluation of automatic story generation. Specifically, we compare the performance of LLMs, with a particular focus on GPT-3, using prompt-based learning, against SOTA models.
We compare generated stories in terms of various automatic evaluation metrics from lexical and semantic matching ones to recently proposed model-based ones.
We follow the best practice in literature to conduct rigorous human evaluations including both crowdworkers from Amazon Mechanical Turk and in-house judges, and assess story quality at a fine-grained level, such as coherence and logicality. 
To summarise, our contributions are:
\begin{itemize}
\item We conduct an empirical comparison between GPT-3 and other SOTA techniques for open-ended story generation on three different corpora that differ in style, register, and length.
\item We test with a wide variety of automatic story evaluation metrics, and find that recent model-based ones work better, consistent with the literature.
\item We conducted experiments using two types of annotators: crowdworkers and in-house judges, to assess the quality of stories on various aspects. The results obtained from two groups are consistent. We release this annotated resource as a testbed for developing new automatic metrics in story generation tasks.\footnote{\url{https://github.com/ZhuohanX/TheNextChapter}}
\item Our experimental findings provide comprehensive evidence that the stories produced by GPT-3 exhibit significant improvement compared to SOTA techniques, and are comparable to stories authored by humans across various aspects.
\item We conduct a preliminary study on story plagiarism and find that GPT-3 tends to (soft) ``plagiarise'' real stories when generating news, 
even though it does not directly copy the source text, 
raising further questions as to what extent GPT-3 recycles stories in its memory rather than generating new narratives.
\end{itemize}

\section{Related Work}
\label{sec:relatedwork}

\paragraph{Story Generation}

\citet{see-etal-2019-massively} 
find that fine-tuned GPT-2 can already generate stories with fluent sentences, but more attentions are needed to incorporate
commonsense and higher-level story planning.
Most works then use PLMs such as GPT-2 or BART as the backbone and incorporate higher level features to aid the generation process. Specifically,
\citet{rashkin-etal-2020-plotmachines, goldfarb-tarrant-etal-2020-content, tan-etal-2021-progressive} construct a storyline to guide the generation process.
\citet{guan-etal-2021-long, yu-etal-2021-sentence, xie-etal-2021-exploring} incorporate inter-sentence relationships such as coherence and discourse relationships into the generation process.
\citet{guan-etal-2020-knowledge, DBLP:journals/corr/abs-2105-01311} explore using external knowledge such as commonsense for story generation.
\citet{xu-etal-2020-megatron, DBLP:conf/aaai/AmmanabroluCBR21} combine storyline planning and commonsense reasoning.

There are also studies which explore the use of GPT-3 for story generation. For example, \citet{clark-etal-2021-thats} conducts a Turing test between GPT-3 generated and human-written stories and \citet{lucy-bamman-2021-gender} probe for gender and representation bias in GPT-3 generated stories. These studies, however, do not provide a systematic evaluation that assesses GPT-3 against the SOTA story generation models.

\paragraph{Story Evaluation}
Automatic story evaluation is admittedly a challenging task, and
the lack of standardized evaluation metrics has somewhat impeded progress of story generation \citep{guan-etal-2021-openmeva}. Human evaluation is usually considered as the gold standard for story quality evaluation, but it is expensive and time-consuming \citep{guan-huang-2020-union} and it can not capture diversity \citep{hashimoto-etal-2017-joint}.
Subsequently, several automated evaluation metrics are introduced as alternative measures to evaluate the quality (the degree of readability) and diversity (the extent of variation) of the generated stories. 
For quality, most metrics measure lexical overlap between strings \citep{papineni-etal-2002-bleu, lin-2004-rouge, tan-etal-2021-progressive} or semantic similarity by comparing embedding of models \citep{zhao-etal-2019-moverscore, DBLP:conf/iclr/ZhangKWWA20} between generated stories and their
human references. 
Recently, learning \citep{sellam-etal-2020-bleurt} and generation \citep{DBLP:conf/nips/YuanNL21} 
based methods are explored and they are based on pre-trained language models such as BERT \citep{devlin-etal-2019-bert} and BART.
Nevertheless, these evaluation metrics are limited in that they provide a single score to indicate the overall quality of the story, and few metrics are specifically designed to assess specific aspects such as logicality (the adherence to commonsense) or interestingness (the level of reader engagement) \citep{DBLP:conf/coling/ChhunCSC22}.



\section{Experimental Setup}

\subsection{Story Generation Models}

To ensure a comprehensive comparison, we conducted extensive experiments involving GPT-3 and a wide range of SOTA story generation models.

In our experiments, we utilized the largest initial version of GPT-3, namely text-davinci-001, which was initially introduced in June 2020 and comprises 175B parameters. It is perhaps worth noting that this model was considered the most powerful at the time of our experiment (March 2022), although subsequent models like GPT-4 have since been released, boasting even greater capabilities. As such, the results we report here can be interpreted as a ``lower bound'' of LLM's story generation performance.
To adapt GPT-3 to the story domain without explicit fine-tuning, we employed a prompt-based learning approach. We selected a small number of stories, typically 2 or 3, to serve as exemplars for GPT-3 in the target domain.

For SOTA story generation models, we use 
1) knowledge enhanced based models: \textbf{KGGPT2} \citep{guan-etal-2020-knowledge} and \textbf{HINT} \citep{guan-etal-2021-long}; 
2) storyline planning based model: \textbf{PROGEN} \citep{tan-etal-2021-progressive}; 
and 3) \textbf{MTCL} \citep{xu-etal-2020-megatron} that combines both storyline planning and commonsense reasoning.
We also fine-tune \textbf{BART} as an additional baseline. For consistency, all models use  
nucleus sampling \citep{DBLP:conf/iclr/HoltzmanBDFC20} with $p = 0.95$ as the decoding method.
We summarise these models in \autoref{table:modeldetails}, and more details can be found in \autoref{appendix: modeldetails}. 

\begin{table*}[t]
\centering
\small
\begin{tabular}{p{1.5cm} p{1.6cm} p{1cm}p{7cm}p{2cm}} 
\toprule
\textbf{Model} & \textbf{Backbone} &  \textbf{Size} & \textbf{Method} & \textbf{Story Datasets}  \\
\midrule
GPT-3 & text-davinci-001 & 175B & Prompt-based learning with several examples from the story dataset (3 for ROC and WP and 2 for CNN) & ROC, WP, CNN \\ 
\midrule
 KGGPT2 & GPT-2 small & 124M & Fine-tuned on commonsense data before more fine-tuning with auxiliary
classification tasks & ROC \\ 
\midrule
PROGEN & BART large & 400M & Three-stage generation where at each stage a fine-tuned BART generates stories based on word importance in the story  datasets & ROC, WP, CNN \\
\midrule
 MTCL & GPT-2 small BERT large & 124M 336M & (1) a GPT-2 model to generate keywords; (2) a BERT model to rank retrieved knowledge triples; and (3) a second GPT-2 model that takes top-ranked knowledge triples and context as input for story generation  & ROC\\ 
\midrule
HINT & BART base & 140M &  BART is first fine-tuned on BookCorpus with additional objectives to learn internal structure in a story and then further fine-tuned on the story  datasets & ROC, WP \\
\midrule
BART & BART large & 400M & Baseline model that is fine-tuned on the story datasets using a standard language modelling objective & ROC, WP, CNN \\
\bottomrule
\end{tabular}
\caption{The backbone (``Backbone'') of the story generation models and their number of parameters (``Size''). ``Story Datasets'' indicates which datasets are used to generate stories for a particular model. KGGPT2 and MTCL stories are obtained from the original authors; for PROGEN and HINT we re-run the implementation provided by the authors.}
\label{table:modeldetails}
\end{table*}

\subsection{Story Datasets}

The most popular story dataset is
ROCStories (ROC) \citep{mostafazadeh-etal-2016-story}, which is composed of short commonsense stories and is used by most story generation works.
There are also more difficult and longer story datasets, such as WritingPrompts (WP) \citep{fan-etal-2018-hierarchical} 
and CNN News (CNN) \citep{DBLP:conf/nips/HermannKGEKSB15} which are composed of fictional and news stories (two different domains).
In our experimental setup, we utilized all three datasets. The ROC dataset was used to evaluate the generation of short stories comprising 5 sentences. The WP dataset was employed to assess medium-length stories, which were trimmed down to 10 sentences. Lastly, the CNN dataset was utilized to evaluate the generation of long stories, and each story have around 20 sentences. For further details about these datasets, please refer to \autoref{appendix:datasetsdetails}.

Whenever possible we evaluate all models on each story dataset.
However, this is sometimes infeasible because some models are designed to work on 
a particular dataset and thus cannot be adapted to other datasets easily. 
Moreover, we focus on {conditional story generation} in this work, this means there is some \textit{context} upon which we generate the stories (details below). 

\paragraph{ROC}
We evaluate all models in this dataset. The context we use to generate stories is the first sentence, 
and so the models are trained to generate the last 4 sentences.
Evaluation results are computed over 800 generated stories using randomly sampled leading sentences from the test partition.


\paragraph{WP}
We assess HINT, PROGEN, 
GPT-3 and BART on this dataset. The context is a short paragraph (``prompt'') that describes the idea of the story.
We randomly sample 1000 prompts from the test partition for automatic evaluation. 

\paragraph{CNN}
We only run GPT-3, BART, PROGEN on CNN, 
as HINT is developed for ROC and WP originally
and it does not work well when applied to CNN. Stories of CNN are generated conditioned on the news titles.
We randomly sample 600 titles from the test partition for automatic evaluation.




\section{Automatic Evaluation}
\subsection{Evaluation Metrics}
We use two types of automatic evaluation metrics: 
1) reference-based metrics, where we compare the generated stories to human reference stories based on the same conditioning context; and 
2) reference-free metrics, where we assess the quality of the stories directly.
\subsubsection{Reference-based Metrics}

Most reference-based metrics measure the lexical or semantic closeness between generated stories and their human references.
We experiment with metrics based on string based matching (CBL, MSJ) and embedding based matching (BES) and a learning based metric (BRT), to assess the quality of generated stories.
We also use a recall based metric (BBL) to assess the diversity of generated stories.
Specifically, \textbf{Corpus BLEU (CBL)}
computes the average BLEU scores \citep{papineni-etal-2002-bleu} 
for each generated story against all human references \citep{DBLP:conf/iclr/CacciaCFLPC20, xie-etal-2021-exploring}.
\textbf{MS-Jaccard (MSJ)} 
measures lexical overlap by computing the n-gram overlap between generated and referenced stories using the Jaccard index \citep{alihosseini-etal-2019-jointly}.
\textbf{BERTScore (BES)}
measures the maximum similarity of each token's contextual embedding between generated and referenced stories \citep{DBLP:conf/iclr/ZhangKWWA20}.
\textbf{BLEURT (BRT)}
is trained on synthetic data to predict a similarity score between generated and referenced stories \citep{sellam-etal-2020-bleurt}.
\textbf{Backward BLEU (BBL)}
computes the coverage of n-grams in the reference stories against the set of generated stories \citep{DBLP:conf/ijcai/ShiCQH18}.\footnote{We use BLEU4 for CBL and BBL; 4-grams overlap for MSJ; roberta-large model for BES; bert-base-128 for BRT.}

\subsubsection{Reference-free Metrics}

Reference-free metrics evaluate generated stories without comparing them to their human-authored references.
We experiment with diversity metrics based on intra-story (D-3, LR-n) and inter-story diversity (SBL).
We also compute negative log-likelihood from BART of a story conditioned on the context (BAS) for relatedness,
and story length in terms of words (LEN) for complexity.

Specifically, \textbf{Lexical Repetition (LR-n)}
computes the average percentage of 4-grams appearing
at least $n$ times in  the generated stories \citep{shao-etal-2019-long}. 
\textbf{Distinct-3 (D-3)} 
computes the average ratio of distinct 3-grams to all 3-grams \citep{li-etal-2016-diversity}.
\textbf{Self-BLEU (SBL)}
measures inter-story diversity that computes the average BLEU score of each generated story using all generated stories as reference \citep{DBLP:conf/sigir/ZhuLZGZWY18}.
\textbf{BARTScore (BAS)}
computes generative likelihood of a story conditioned on the context 
(i.e., leading sentence for ROC, prompt for WP and title for CNN)
to measure the extent to which a generated story relates to its condition \citep{DBLP:conf/nips/YuanNL21}.\footnote{We set $n=$ 3/8/8 for ROC, WP and CNN respectively and use BLEU4 for SBL. We use the ``PARA'' version of BART and direction as ``from source to hypothesis''.
}
\textbf{Length (LEN)}
measures the average length of the generated stories, which is used as 
a rough indicator of generation complexity.

\subsection{Results}
\label{sec:automatic-results}

\autoref{table:re} and \autoref{table:ure} present the reference-based and reference-free evaluation results, respectively.
At a glance, these metrics do not appear to agree with each other 
even though some of them are designed to evaluate the same aspect (e.g.,\ the best model in terms of fluency/coherence or diversity is different depending on the metric). Overall, GPT-3 seems to have weaker performance than most of other models in terms of quality (CBL and MSJ) and diversity (BBL, SBL, D-3 and LR-n) metrics.

\begin{table}[t]
\centering
\small
\begin{tabular}{p{1ex}lccccc}
\toprule
 &\multirow{3}{3em}{\textbf{Model}} & \multicolumn{4}{c}{\textbf{Flu./Coh.}} & \textbf{Div.}\\
\cmidrule(lr{0.25em}){3-6}\cmidrule(l{0.25em}r){7-7} 
 & & {\textbf{CBL}}  &  \textbf{MSJ}  & \textbf{BES}  & \textbf{BRT}  & \textbf{BBL}  \\
 & & {$\Uparrow$} & $\Uparrow$ & $\Uparrow$ & $\Downarrow$ & $\Uparrow$ \\
\midrule
 \multirow{7}{3em}{\rotatebox{90}{\textbf{ROC}}} 
 & GPT-3 & 27.2 & 11.6 & 86.6 & \textbf{8.6} & 24.0 \\ 
 & KGGPT2 & 33.5 & \textbf{15.0} & \textbf{87.0} & 9.5 & \textbf{25.6} \\ 
 & PROGEN3 & 26.6 & 14.6 & 86.7 & 9.7 & 25.0 \\ 
 & MTCL & 31.4 & 14.2 & 86.9 & 9.7 & 24.0 \\
 & HINT & \textbf{39.6} & 13.7 & \textbf{87.0} & \textbf{8.6} & 24.6 \\
 & BART & 27.5 & 14.7 & 86.8 & 9.5 & 25.1 \\
\midrule
\multirow{5}{3em}{\rotatebox{90}{\textbf{WP}}} 
 & GPT-3 & 28.6 & 12.3 & \textbf{81.6} & \textbf{11.7} & 24.4 \\ 
 & PROGEN3 & 32.3  & \textbf{16.4} & 81.4 & 13.3 & \textbf{27.6} \\
 & HINT & \textbf{45.5}  & 12.8 & 80.8 & 12.1 &  23.7 \\ 
 & BART & 32.6 & 16.2 & 81.4 & 13.0 & 27.2 \\
 \midrule
 \multirow{3}{3em}{\rotatebox{90}{\textbf{CNN}}} 
 & GPT-3 & \textbf{33.2} & 11.0 & \textbf{83.5} & \textbf{7.5} & 19.8 \\ 
 & PROGEN3 & 29.6 & \textbf{14.8} & 82.2 & 9.3 & \textbf{26.2} \\
 & BART & 29.1 & 14.7 & 82.2 & 9.8 & 25.7 \\
 \bottomrule
\end{tabular}
\caption{Reference-based Evaluation Results.
CBL, MSJ, BES and BRT evaluate the closeness between the generated stories and
the whole test reference data as an indicator of general fluency (Flu.) and coherence (Coh.).
BBL focus on the recall of generated stories as an indicator of diversity (Div.).
$\Uparrow$: higher is better; $\Downarrow$: lower is better.
BRT values are negated here.}
\label{table:re}
\end{table}

\begin{table}[t]
\centering
\small
\begin{tabular}{p{1ex}lccccc}
\toprule
 &\multirow{3}{3em}{\textbf{Model}} & \multicolumn{3}{c}{\textbf{Div.}} & \textbf{Rel.} & \textbf{Com.}\\
\cmidrule(lr{0.25em}){3-5}\cmidrule(lr{0.25em}){6-6}\cmidrule(l{0.25em}r){7-7} 
 & & \textbf{SBL}  &  \textbf{D-3}  & \textbf{LR-n} & \textbf{BAS}  & \textbf{LEN} \\
  & & {$\Downarrow$} & $\Uparrow$ & $\Downarrow$ & $\Downarrow$ & $\Uparrow$ \\
\midrule
 \multirow{8}{3em}{\rotatebox{90}{\textbf{ROC}}} 
 & GPT-3 & 38.5 & 67.7 & 39.1 & \textbf{4.2} & 47.3 \\ 
 & KGGPT2 & 41.9 & 67.2 & 51.9 & 4.6 & 38.4 \\ 
 & PROGEN3 & \textbf{30.0} & 76.9 & 39.5 & 5.0 & 40.9 \\ 
 & MTCL & 39.4 & 69.6 & 44.4 & 4.9 & \textbf{49.7} \\
 & HINT & 55.1 & 54.3 & 68.1 & 4.3 & 35.8 \\
 & BART & 30.5 & 77.4 & 37.8 & 5.0 & 40.6 \\
 & human & 33.1 & \textbf{80.2} & \textbf{35.8} & 5.2 & 40.3 \\
\midrule
\multirow{6}{3em}{\rotatebox{90}{\textbf{WP}}}
 & GPT-3 & 37.5 & 69.6 & 9.7 & 4.3 & 120.6 \\ 
 & PROGEN3 & 35.2 & 77.2 & 2.6 & 5.4 & 136.9 \\
 & HINT & 64.1 & 33.9 & 67.4 & \textbf{4.1} & 119.0 \\ 
 & BART & 35.3 & 77.5 & 1.6 & 5.4 & 129.2 \\
 & human & \textbf{27.1} & \textbf{83.7} & \textbf{1.5} & 5.7 & \textbf{150.0} \\
 \midrule
 \multirow{5}{3em}{\rotatebox{90}{\textbf{CNN}}} 
 & GPT-3 & \textbf{26.5} & 82.9 & 9.8 & \textbf{4.4} & 147.3 \\ 
 & PROGEN3 & 28.9 & 82.3 & 2.3 & 5.2 & 395.8 \\
 & BART & 27.9 & 83.2 & \textbf{0.8} & 5.2 & 374.1 \\
 & human & 27.3 & \textbf{83.8} & 6.3 & 5.4 & \textbf{498.6} \\
 \bottomrule
\end{tabular}
\caption{Reference-free Evaluation Results.
SBL measures inter-story diversity by assessing differences between different stories while
D-3 and LR-n (3 for ROC, 8 for WP and CNN) focus on repetition n-grams within the same story.
We also include LEN as an indicator of story complexity (Com.).
We compute BAS of story given condition for story relatedness (Rel.).
}
\label{table:ure}
\end{table}


However, when we look at recent model-based metrics (BERTScore, BLEURT and BARTScore), GPT-3 appears to be a much better model (a finding we will return to when we look at human evaluation results).
Interestingly, we notice that human written stories have very poor performance in terms of BARTScore (BAS).
We suspect BARTScore may exhibit a bias towards machine-generated stories, as the metric primarily evaluates quality based on the generative likelihood of a sequence. Machine-generated stories are specifically designed to maximize this likelihood, while human-authored stories often incorporate distinct elements, such as surprising or creative word choices \citep{DBLP:conf/iclr/HoltzmanBDFC20}.
In general, all models are capable of generating stories of the appropriate length, except for GPT-3 in the CNN dataset. GPT-3 in the CNN dataset exhibits difficulty in generating stories longer than 150 words, whereas human-written stories typically consist of around 500 words on average. Considering the overall assessment using various automatic metrics, there is no single winner  that consistently outperforms other models.

\section{Human Evaluation}

To obtain a comprehensive assessment of the generated stories, we recruit human annotators to evaluate their quality. In order to gain insights into consistency, we employed both crowdsourced workers and in-house annotators. This approach allows us to gather diverse perspectives and obtain a more nuanced understanding of the story quality.

\subsection{Crowdsource Annotation}
\label{subsec:crowdsourceannotations}
We first collect human judgements using the Amazon Mechanical Turk (AMT) platform.\footnote{\url{https://requester.mturk.com/}}
Following the approach suggested by \citet{karpinska-etal-2021-perils}, we assessed four aspects, namely fluency, coherence, relatedness, and interestingness. Additionally, we introduced a new aspect called logicality, which assesses the extent to which the story complies with commonsense. Each of these five aspects is evaluated on an ordinal scale ranging from 1 (worst) to 5 (best). 
We randomly sample 20 conditional contexts (e.g.,\ titles) from each dataset 
and collect stories generated by all models for human evaluation.
Each story (including human-written one) is judged by 3 annotators, and so we have annotations for 320 stories in total (140/100/80 for ROC, WP and CNN, respectively). 
Amazon 
Qualification requirements on AMT and question details can be found in \autoref{appendix: mturksetting}.
Quality control details can be found in \autoref{appendix: pilotstudy}.


\begin{table}[t]
\centering
\small
\begin{tabular}{p{1ex}p{8ex}p{4.5ex}p{4.5ex}p{4.5ex}p{4.5ex}p{4.5ex}}
\toprule
& \textbf{Model} & \textbf{Flu.} &  \textbf{Coh.} & \textbf{Rel.} & \textbf{Log.} & \textbf{Int.}  \\
\midrule
\multirow{7}{3em}{\rotatebox{90}{\textbf{ROC}}} & GPT-3 & $\mathbf{4.40}$ & $4.43$ & $4.37$ & $4.37$ & $3.57$ \\ 
 & KGGPT2 & $3.90^{*}$ & $3.48^{*}$ & $3.53^{*}$ & $3.00^{*}$ & $2.62^{*}$ \\
 & PROGEN3 & $3.88^{*}$ & $3.45^{*}$ & $3.37^{*}$ & $2.95^{*}$ & $2.57^{*}$ \\
 & MTCL & $3.55^{*}$ & $3.12^{*}$ & $3.18^{*}$ & $2.73^{*}$ & $2.42^{*}$ \\
 & HINT & $3.90^{*}$ & $3.27^{*}$ & $3.33^{*}$ & $3.12^{*}$ & $2.58^{*}$ \\
 & BART & $3.92^{*}$ & $3.38^{*}$ & $3.48^{*}$ & $3.03^{*}$ & $2.60^{*}$ \\
 & human & $4.22$ & $\mathbf{4.58}$ & $\mathbf{4.42}$ & $\mathbf{4.48}$ & $\mathbf{3.77}$ \\
 \midrule
 \multirow{5}{3em}{\rotatebox{90}{\textbf{WP}}} & GPT-3 & \textbf{4.37} & \textbf{4.67} & \textbf{4.28} & \textbf{4.48} & \textbf{3.47} \\ 
 & PROGEN3 & $3.45^{*}$ & $3.08^{*}$ & $2.35^{*}$ & $2.57^{*}$ & $1.98^{*}$ \\
 & HINT & $3.32^{*}$ & $2.63^{*}$ & $2.02^{*}$ & $2.25^{*}$ & $1.77^{*}$ \\
 & BART & $3.42^{*}$ & $2.73^{*}$ & $2.08^{*}$ & $2.27^{*}$ & $1.87^{*}$ \\
 & human & $4.13^{*}$ & $4.22^{*}$ & $3.05^{*}$ & $3.75^{*}$ & $2.97^{*}$ \\
 \midrule
\multirow{4}{3em}{\rotatebox{90}{\textbf{CNN}}} & GPT-3 & $\mathbf{4.22}$ & $\mathbf{4.52}$ & $\mathbf{4.58}$ & $\mathbf{4.60}$ & $3.20$ \\ 
 & PROGEN3 & $3.63^{*}$ & $3.32^{*}$ & $3.30^{*}$ & $3.22^{*}$ & $2.28^{*}$ \\
 & BART & $3.58^{*}$ & $3.37^{*}$ & $3.30^{*}$ & $3.27^{*}$ & $2.17^{*}$ \\
 & human & 4.10 & $4.10^{*}$ & $4.23^{*}$ & $4.18^{*}$ & $\mathbf{3.72^{*}}$ \\
\bottomrule
\end{tabular}
\caption{Crowdsource Human Evaluation Results. 
We calculate the average score of models for each aspect:
fluency (Flu.), coherence (Coh.), relatedness (Rel.), logicality (Log.) and
interestingness (Int.).
Model scores that are marked with $*$ indicate the performance difference
between the model and GPT-3 is significant.
}
\label{table:he}
\end{table}

\label{sec:humanevaluationresults}
\autoref{table:he} presents the human evaluation results.
Overall, GPT-3 generates stories that are consistently of higher quality than
those generated by other SOTA models. 
To understand whether the difference is significant,
we perform a paired t-test by comparing GPT-3 to other models (including human)
and find that in most cases 
these results are significant with $p$-value < 0.05 (`*' in the table).
Compared with human authors, GPT-3 appears that it is generating stories that are just as good as (ROC) or better than (WP and CNN) human authors, confirming the findings of \citet{clark-etal-2021-thats}. 
For WP, in particular, human stories are trimmed to the first 10 sentences (data pre-processing for training the story generation models).
This abruptly shortens the stories so they might not provide a proper conclusion, 
and inevitably are penalised (see examples in \autoref{sec:GPT-3 Stories vs Human References}).
For CNN, GPT-3 appears to be ``plagiarising'' real stories, where many story elements are not a product of creative generation but details copied from real news stories (\autoref{sec:manualcheck}). 
Another reason could be that GPT-3 stories are much shorter than 
those generated by other models and human authors (150 vs.\ 300-400 words; \autoref{table:ure}), 
which makes them easier to read and thus leads to better scores. 
Note that this is a downside of GPT-3 where it is difficult to get it to generate long stories (\autoref{sec:discussion}).

\begin{table}[t]
\centering
\small
\begin{tabular}{p{1ex}p{8ex}p{4.5ex}p{4.5ex}p{4.5ex}p{4.5ex}p{4.5ex}}
\toprule
& \textbf{Model} & \textbf{Flu.} &  \textbf{Coh.} & \textbf{Rel.} & \textbf{Log.} & \textbf{Int.}  \\
\midrule
\multirow{7}{3em}{\rotatebox{90}{\textbf{ROC}}} & GPT-3 & $\mathbf{4.78}$ & $\mathbf{4.73}$ & $\mathbf{4.50}$ & $\mathbf{4.82}$ & $\mathbf{3.37}$ \\ 
 & KGGPT2 & $4.52^{*}$ & $3.67^{*}$ & $3.57^{*}$ & $3.47^{*}$ & $2.50^{*}$ \\
 & PROGEN3 & $4.27^{*}$ & $3.47^{*}$ & $3.78^{*}$ & $3.23^{*}$ & $2.48^{*}$ \\
 & MTCL & $4.27^{*}$ & $3.27^{*}$ & $3.45^{*}$ & $3.15^{*}$ & $2.37^{*}$ \\
 & HINT & $4.38^{*}$ & $4.03^{*}$ & $3.38^{*}$ & $3.70^{*}$ & $2.38^{*}$ \\
 & BART & $4.37^{*}$ & $3.95^{*}$ & $3.85^{*}$ & $3.53^{*}$ & $2.70^{*}$ \\
 & human & $4.52^{*}$ & $4.38^{*}$ & $4.22$ & $4.32{*}$ & $3.18$ \\
 \midrule
 \multirow{5}{3em}{\rotatebox{90}{\textbf{WP}}} & GPT-3 & \textbf{4.57} & \textbf{4.65} & \textbf{4.08} & \textbf{4.22} & \textbf{3.82} \\ 
 & PROGEN3 & $3.55^{*}$ & $3.03^{*}$ & $2.23^{*}$ & $2.57^{*}$ & $2.45^{*}$ \\
 & HINT & $3.60^{*}$ & $2.72^{*}$ & $2.07^{*}$ & $2.68^{*}$ & $2.08^{*}$ \\
 & BART & $3.45^{*}$ & $2.77^{*}$ & $2.08^{*}$ & $2.38^{*}$ & $2.30^{*}$ \\
 & human & $4.05^{*}$ & $4.07^{*}$ & $3.73$ & $3.87^{*}$ & $3.78$ \\
 \midrule
\multirow{4}{3em}{\rotatebox{90}{\textbf{CNN}}} & GPT-3 & $\mathbf{4.50}$ & $\mathbf{4.33}$ & $\mathbf{4.48}$ & $\mathbf{4.40}$ & $\mathbf{3.45}$ \\ 
 & PROGEN3 & $3.80^{*}$ & $3.45^{*}$ & $3.63^{*}$ & $3.45^{*}$ & $2.52^{*}$ \\
 & BART & $3.73^{*}$ & $3.25^{*}$ & $3.58^{*}$ & $3.32^{*}$ & $2.57^{*}$ \\
 & human & $4.22^{*}$ & $4.00^{*}$ & $4.35$ & $4.13^{*}$ & $3.22$ \\
\bottomrule
\end{tabular}
\caption{In-house Human Evaluation Results. 
}
\label{table:he-inhouse}
\end{table}

When considering the various aspects for SOTA models, including KGGPT2, PROGEN3, MTCL, HINT, and BART, these models exhibit strong performance in terms of fluency, with scores consistently exceeding 3.5 in most cases. This indicates that the models can generate sentences that are natural and fluent.
However, coherence performance differs depending on the dataset. Most models perform well on the ROC and CNN datasets, while they tend to struggle on WP, with coherence scores falling below 3.1.
The observation that these models struggle with shorter WP stories compared to longer CNN stories might be because the PLMs that they are built on are mostly trained on web data which contains plenty of news articles.
For relatedness, logicality and interestingness, we see a similar trend where the models perform best in ROC and worst in WP. 
We also observe a consistent decrease in performance from relatedness to logicality and interestingness,
suggesting that the models particularly struggle to generate interesting and sensible stories.
Interestingness is perhaps the most difficult aspect to optimise, 
as it is difficult to define what makes a narrative interesting.

\subsection{In-house Annotation}
\label{subsec:inhouseannotations}
We next recruit university volunteers to collect in-house judgements.\footnote{Demographically, 14 are PhD students and 1 is university staff; all of them are proficient in English.}
We ask them to evaluate the same 5 aspects using the same scale.
We sample 20 \textit{disjoint} conditional contexts from each dataset for story generation here, as we are interested to test the robustness of our previous findings (with different workers and set of stories). 
As with crowdsource annotation, each story is also judged by 3 annotators.
Details of the agreement between annotators can be found in \autoref{appendix:IAA-inhouse}.

\autoref{table:he-inhouse} presents the scores of story quality from in-house annotators. Interestingly, the \textit{magnitude} of the in-house scores are generally somewhat higher than the crowdworker scores (across all metrics and datasets and models). We hypothesise that this may be because our in-house workers are more ``tolerant'' to mistakes as they have been exposed to machine-generated text more compared to crowdworkers. That said, the overall findings are  consistent between the two groups of annotators:
1) GPT-3 is the best story generation model and outperforms both SOTA models and human stories;
2) The SOTA models do well in fluency, but poorly in most other aspects (interestingness worst); and
3) The SOTA models face notable challenges specifically in WP, as evidenced by their poor coherence, relatedness, logicality scores compared to other domains.



When comparing the results of automatic metrics (\autoref{sec:automatic-results}) to the human evaluation results, a notable discrepancy emerges, leading to a different conclusion regarding the performance of GPT-3 and the identification of a clear ``best'' story generation model. That said, if we consider only model-based metrics such as BERTScore, BLEURT for fluency/coherence, and BARTScore for relatedness, a more aligned conclusion can be drawn, suggesting these metrics may be more reliable (though the trend is still less conclusive compared to human evaluation results). This observation is consistent with recent literature, which highlights the better correlation of modern model-based metrics with human evaluations \citep{llmevaluation-acl23, decompeval-acl23, deltascore-arxiv23}.

%


\section{Plagiarism}
\label{sec:plagiarismcheck}

Considering the strong performance of GPT-3 on story generation,
we next provide a preliminary investigation 
to understand the extent to which GPT-3 copies from its training data.

\paragraph{IThenticate}
We use iThenticate\footnote{\url{https://www.ithenticate.com}} --- a professional plagiarism detection software that has comprehensive coverage over online articles ---  
to assess how much GPT-3 plagiarises. 
We include only the generated content (without the condition) 
when checking for plagiarism.
Results show that there is no strong plagiarism: 
similarity score for ROC, WP and CNN is 4\%,
3\% and 14\% respectively. 
This is consistent with findings from \citet{DBLP:journals/corr/abs-2111-09509} that the language models do not simply memorize, but combine familiar parts into novel way.
However, iThenticate looks for lexical overlap to identify plagiarism. 
A more subtle form of plagiarism is one that copies the ideas without parroting the words \citep{DBLP:journals/corr/abs-2203-07618}, motivating us to perform manual check.


\paragraph{Manual Check}
\label{sec:manualcheck}

We randomly sample 10 GPT-3 generated stories for WP and CNN respectively and manually check whether 
they copy from web articles by searching for keywords, named entities and long phrases.
For WP, in most instances we do not find web articles that resemble our stories, 
except for one instance where it copies a story from Wikipedia (noting that it is not a direct word-to-word copy).
For CNN, we find 7 out of 10 instances are based on real stories from news articles, where most details (e.g.,\ names, incident origins, sentence penalty) are faithful to the original stories. 
These results highlight that GPT-3 can recall details from training data, 
calling into question how much of its generation is creative. 
More details can be found in \autoref{sec:Soft Plagiarism Examples}.


\section{Discussion}
\label{sec:discussion}

\begin{table*}[t]
\centering
\small
\begin{tabular}{p{1.3cm} p{13.5cm}} 
\toprule
\textbf{Model} & \textbf{Story}  \\
\midrule
GPT-3 & `` yes , lord cthulhu ? '' you answer the phone . `` i need you to pick me up from the airport . '' `` of course , my lord . when will you be arriving ? '' `` in about an hour . i 'll text you the flight number . '' you hang up and start getting ready . you know cthulhu is a demanding deity , but you do n't mind . you enjoy serving him . \\
\midrule
 PROGEN3 & **but everything 's different ! i miss my child in there ...** *what are you talking about ! you have to be mad . you must have misheard ! ! you come and get your mommy ... '' momma always grabbed him before they climbed into the bathtub . after that happened they were always late . * bam . * wait , it 's never like this . ctlulhu always puts up a fight . i repeat : it 's never like this ! it 's always for the umpteenth time , now or never . *the phone rings . i answer it and ctlulhu 's voice booms out .\\ 
\midrule
 HINT & `` sir , i 'm afraid i 'm late for work. `` `` i 'm sorry sir , `` the technician said , `` but i have to get my cell phone back. `` `` sir , you have to pick up this cell. `` `` yes sir , i do. `` `` and how are you going to find it ? `` `` i 'm just going to have to pick it up. `` `` well , sir , `` the technician said , `` it 's just that i do n't have the time to pick it up. `` `` why not ? `` `` sir , `` the technician said , `` i do n't have the time. `` `` i 'm just going to make sure it 's picked up. `` \\ 
\midrule
BART & “ ah it ’ s here . i thought you wanted to leave , a phone call . * ” my coworkers looked at me confused . “ i wasn ’ t in the mood to go running around scared . but it was important . you need to pick up on your cravings . the kid-ta-bye , gabriel , payliver , cilia de los angeles . and four people ’ s fault. ” he was right about the fifth , at least my attention was fixed on it . three girls were outside screaming in the street , and a family of four was picking up one of them . the two men yelled and approached as they walked to the mailbox , visibly distressed . \\
\bottomrule
\end{tabular}
\caption{Stories generated by GPT-3, HINT, BART and PROGEN3 given the same title ``cthulhu calls your cell , he needs to be picked up .''}
\label{table:demostories}
\end{table*}

In this paper, we only experiment with a simple prompt-based learning approach to adapt GPT-3 to the different story datasets and one can argue more prompt engineering can be explored \citep{liu-etal-2022-makes, lu-etal-2022-fantastically, mishra-etal-2022-reframing}. 
We contend, however, that our goal here is not to propose a better prompt-based method to do story generation with GPT-3, 
and ultimately even with our simple approach we find that GPT-3 generates high quality stories, 
suggesting that with additional prompt engineering it can do better.
We next discuss \textit{qualitatively} some of the strengths and drawbacks of GPT-3 in story generation.

\subsection{Strengths}

Unlike other conditional generation tasks such as machine translation or summarisation, where
the input context is rich in information and the goal is to translate or compress the input information,
story generation works in the reverse manner where the model needs to ``hallucinate'' new information and details given a succinct context.
This means that in order to do the task well, having strong world knowledge is important. Reading some of the GPT-3 stories, we observe GPT-3's advantage in this, particularly in the WP dataset where some of the prompts require niche knowledge about characters. In \autoref{table:demostories} we show an example in WP where the prompt is \textit{cthulhu calls your cell , he needs to be picked up}, where \textit{cthulhu} a fictional cosmic entity, and only GPT-3 is able to produce a coherent story and the SOTA models struggle.


\subsection{Drawbacks}

Even though GPT-3 demonstrates excellent generative capability and outperforms SOTA models significantly,
we still find GPT-3 has many generation errors that can be improved.

\paragraph{Story length}
GPT-3 has a parameter to control the maximum number of generated tokens but does not provides a way to control the minimum number of tokens.
As one can see from \autoref{table:ure}, 
GPT-3 can not generate stories longer than 150 words for CNN,
even though the prompts have long stories. 
We also attempted to encourage longer stories by adding specific instructions as part of the prompt of GPT-3, but this did not work.
\paragraph{Null generation}
Occasionally GPT-3 decides to generate no output. 
This is usually not an issue, since this can be solved by forcing it to generate again, although it is unclear why this occurs.
\paragraph{Direct copy}
Besides the soft plagiarism issue (\autoref{sec:manualcheck}), 
GPT-3 does occasionally copy long chunks of text, 
e.g.,\ the title or prompt in the story.
\paragraph{Multilingual} GPT-3 sometimes generates stories in languages other than English, 
despite the given prompts always being in English.
In terms of statistics, out of 1000 generations we find 14 non-English stories (5 Chinese, 4 German, 1 Japanese, 1 French, 1 Russian, 1 Norwegian Nynorsk and 1 mixture of Chinese and English).
Interestingly, in most of these cases the stories are related to the condition (even though in different languages) although sometimes 
we observe the outputs are direct translation of the prompt and not a creative story.
\paragraph{Tokenisation issue}
GPT-3 generations occasionally feature ``sticky'' words where there are missing white spaces (e.g.,\ \textit{understand.With} and \textit{timewhen}). 
We suspect this is due to Byte-Pair Encoding of GPT-3 where white spaces are ``glued'' 
to each subword and so every subword has two versions 
(one with the white space and one without). 
This issue arises when GPT-3 generates using a subword without the white space suffix.

\paragraph{Expletives}
GPT-3 would occasionally generate stories with expletives. 
Interestingly, it would sometimes self-censor them (e.g.,\ \textit{b****}).



\section{Conclusion}
We present an extensive comparison of GPT-3 with SOTA models for story generation, and found
that stories generated by GPT-3
are substantially better than SOTA models on multiple aspects and 
even rival human authors.
The findings of this study indicate that we have entered a new chapter in story generation with LLMs. Future research is likely to concentrate on prompt-engineering LLMs to achieve enhanced customization, such as varying their style and length, further advancing the capabilities of story generation models.
In terms of evaluation metrics, our work: 1) reveals a weak correlation between automatic lexical-based evaluation metrics and human evaluation, and that recently proposed model-based metrics appear to more reliable; and 2) contributes a new test bed for metric development, through the release of a dataset that contains story quality annotations by two groups of judges.
In spite of the positive results of GPT-3 in story generation, we discuss some of its issues, the chief one being that it has a tendency to reproduce details or plots from its memories, raising foundational questions about its generation creativity.

\section*{Limitations}
As observed by \citet{mishra-etal-2022-reframing}, engineering appropriate prompts can significantly influence the performance of language models. In our current study, we randomly sample a few training examples as demonstrations for GPT-3 (in-context learning). However, a more effective approach could involve strategically selecting contextually more relevant examples. 

Although text-davinci-001 was the best model at the time of our experiment, recent advancements in the field have led to the release of more powerful LLMs.
Despite these improved models, we hold the view that they are unlikely to substantially alter the conclusions drawn in this study. The findings strongly suggest that LLMs will remain the dominant approach in story generation in the foreseeable future.
Also, we only explore with GPT-3 in our experiments, and although we think our findings are likely to generalise to other LLMs, this has not been empirirically validated.

Since we started this work in 2022, there has been quite a development in terms of text generation evaluation metrics  \citep{llmevaluation-acl23, decompeval-acl23, deltascore-arxiv23, gptscore-arxiv23, geval-arxiv23}, and some of these uses LLMs themselves. Although we claim that human evaluation remains the gold standard for story generation, it remains to be seen how much these new metrics close the gap. We foresee that the question of circularity, i.e.\ using LLMs to evaluate LLM-generated text, will be the next challenge that the field needs to address.

In our work, we acknowledge that we did not involve domain experts (e.g., story writers) for a more specialized assessment. It would be intriguing to investigate the potential variations in judgments between lay individuals and expert evaluators in story assessment \citep{llmevaluation-acl23}.
Recent research has indicated that certain practices in the human annotation process, such as the use of Likert scales, have limitations in capturing the true preferences of humans \citep{authenticitygap-emnlp22, rose-acl23}. We contend, however, that the fact that we found consistent results between two different groups of annotators suggest that our findings are likely to be robust.



\section*{Ethics Statement}


All mechanical turk experiments conducted in this paper were approved by 
internal ethics review board from our institution.
(Ethics ID Number: 21961).
Our evaluators were paid based on an estimated US\$14.83 per hour rate.
For each dataset, we estimate the time they would spend and 
vary the payment according to the estimated time.
Each HIT contains 7 stories (5 stories to be evaluated 
and 2 controlled stories to control the evaluation quality on AMT).
We pay US\$2.50 per HIT for ROC, US\$3.50 for WP and US\$4.50 for CNN.

We remind the workers in our consent form that the potential risks about this work,
which they might have to read and evaluate stories with filthy words or
offended storyline and they are welcome to quit the task and we will still pay
them according to the efforts they spend.


\section*{Acknowledgements}
We extend our thanks to the reviewers for their valuable feedback, which has greatly contributed to the improvement of this work. We also appreciate the efforts of our online annotators and colleagues for their helpful annotation contributions. Zhuohan Xie is supported by Melbourne Research Scholarship, and would like to expresses his sincere appreciation to the program.


\bibliography{anthology,custom}
\bibliographystyle{acl_natbib}


\clearpage

\appendix

\section{SOTA Story Models Details}
\label{appendix: modeldetails}

\paragraph{Knowledge Enhanced GPT-2 (KGGPT2)} 
\citet{guan-etal-2020-knowledge} use heuristic rules to translate commonsense triples 
from commonsense knowledge bases (e.g.,\ ConceptNet \citep{speer-havasi-2012-representing} and ATOMIC \citep{DBLP:conf/aaai/SapBABLRRSC19})
into natural language sentences
and fine-tune GPT-2 small using these sentences. 
They also use rules to construct negative samples from the original stories to create ``bad stories'' and
perform additional training to encourage the model to learn representations that can distinguish the original and negative stories on ROC.

\paragraph{Progressive Generation of Long Text (PROGEN)} 
\citet{tan-etal-2021-progressive} divide the story generation process into 
multiple stages where words are generated based on their order of importance (estimated using TF-IDF). In other words, PROGEN does not generate stories in a left to right manner.
They fine-tune BART-large in different stages where the early stages focus on generated keywords and the intermediate stages focus on generating the next set of content words.
We use PROGEN3 in our experiment which has 3 stages where it generates 15\%/25\%/100\% of the story words after each pass.

\paragraph{MEGATRON-CNTRL (MTCL)} 
\citet{xu-etal-2020-megatron} combines commonsense reasoning and storyline planning. 
They first train a keyword predictor with GPT-2 and the predicted keywords are used to retrieve
related knowledge triples from a knowledge base.
They then train a contextual knowledge ranker with BERT to rank the top-$N$ predicted knowledge triples.
A second GPT-2 is trained as a conditional generator that takes both top ranked knowledge triples and other conditioning (e.g.,\ titles) as input when generating stories. Note that the parameters of the two GPT-2 and BERT models are initialised using MEGATRON parameters \citep{DBLP:journals/corr/abs-1909-08053}.

\paragraph{High-Level Representations for Long Text Generation (HINT)}
\citet{guan-etal-2021-long} pre-train BART-base on BookCorpus \citep{DBLP:conf/iccv/ZhuKZSUTF15} 
with additional objectives that capture
 sentence-level similarity and sentence-order to 
learn the internal structure within a story. 
The model is then further fine-tuned on story datasets to generate stories in a particular dataset.

\paragraph{BART} This is a baseline model where we fine-tune BART-large on the story datasets with the standard next word prediction objective.

\section{Datasets Details}
\label{appendix:datasetsdetails}

\paragraph{ROCStories (ROC)}

ROC was developed by \citet{mostafazadeh-etal-2016-story} and it contains
98K commonsense stories of five sentences.
To obtain a more generalised lexicon, 
we follow the delexicalisation process from prior studies \citep{guan-etal-2020-knowledge, 
xu-etal-2020-megatron} where male/female/unknown names are replaced by 
 [MALE]/[FEMALE]/[NEUTRAL] sentinels. For each story, the first (leading) sentence is used as conditioning context, and models are trained to generate the remaining 4 sentences.

\paragraph{WritingPrompts (WP)}

WP consists of 303K human-written stories mined from Reddit's Writing Prompts
forum \citet{fan-etal-2018-hierarchical}.\footnote{\url{https://www.reddit.com/r/WritingPrompts/}} Each story is trimmed to contain only the first 10 sentences (following \citet{guan-etal-2021-long}). For WP, we use the prompt (which is typically a paragraph of text that sets the scene of the story) as conditioning for story generation.

\paragraph{CNN News (CNN)}

CNN News \citep{DBLP:conf/nips/HermannKGEKSB15} is a dataset that contains long news articles with titles.
CNN is a very large dataset, with 311K news articles and highlights.
We sub-sample the standard training, validation and testing splits to produce splits with 10K/5K/1K stories each, respectively, for our experiments. The title of a news story is used as conditioning for story generation.

\section{Amazon Mechanic Turk Setting}
\label{appendix: mturksetting}

\paragraph{Qualification Requirements} 
We set following qualification requirements for our annotators:
1) Their accept rate is greater than or equal to 97\%.
2) Their location is in US.
3) They have to complete more than 1000 HITs.

\paragraph{Questions}

We ask the following questions in our questionnaire.

\begin{enumerate}
    \item Fluency: ``How grammatically correct is the text of the story?''
    \item Coherence: ``How well do the sentences in the story fit together?''
    \item Relatedness: ``How relevant is the story to the title?''
    \item Logicality: ``How much does the story obey commonsense?''
    \item Interestingness: ``How enjoyable do you find the story?''
\end{enumerate}

\section{Amazon Mechanic Turk Pilot Study}


\label{appendix: pilotstudy}
While AMT is convenient to find workers for annotation work,
it can be rather difficult to obtain reliable workers \citep{karpinska-etal-2021-perils, clark-etal-2021-thats}.
One of our workers told us many workers install website plugins to help them
to manage the workflow with AMT so that they can hoard many HITs at the same time.
Therefore, HITs with high payment can easily attract irresponsible workers 
even though previous qualifications are set since most AMT requesters will not bother
to reject work.

Therefore, we set a pilot study to aid us to help reliable workers.
We randomly select 5 stories generated from different models on ROC
and 1 story from the test dataset.
We then train a trigram language model on ROC to mimic the style and
generate 1 story from the trigram model.
All stories have different titles.
We randomly shuffle these 7 stories and the task is to ask people to evaluate
all stories with questions mentioned in \autoref{appendix: mturksetting}
and we will judge the quality of their evaluation based on human and trigram stories.

We invite 7 of our colleagues, which are all from non-English speaking countries
to have a rough idea of the difficulty degree of the task.
We calculate the average score of all quality metrics except 
the interestingness aspect since it is subjective.
On average, our colleagues rank the human story as 4.5 and trigram story as 1.425, 
which shows our task is not hard to distinguish human and trigram stories.
We set a rather lenient standard as ``ranking human story >= 3.5 and trigram story <= 2.0" to
select workers from our pilot study.

We create 100 assignments of the same HIT 
at different times with the qualification mentioned in \autoref{appendix: mturksetting}.
We find running the same pilot study at different times can obtain quite different results
from AMT, which align to the findings in \citet{karpinska-etal-2021-perils}.
Generally, we find that more reliable workers can be found in the evening of Eastern Daylight Time (EDT).
We have 10 out of 100 people pass the pilot study but only 5 people pass it on the day.
It shows the difficulty of obtaining reliable workers on AMT nowadays and the economic importance of running a pilot study before conducting real research.
We grant those reliable workers the customised qualification and
only invite them to our real study,
we also have controlled stories to monitor the quality of workers,
as 2 controlled stories inserted into each HIT.

\section{Amazon Mechanic Turk Issue}

Our human evaluation is conducted over AMT, even though it is convenient and affordable, we find a big disagreement between our annotators.
We first conduct a pilot study to test the capability of annotators to evaluate English stories and only invite workers that pass our proficient English stories reading tests to the evaluation of sampled stories.
We only gave them two examples showing how we assess the example stories but we did not provide detailed English stories evaluation training to our annotators.
We did not have a main annotators that can provide a standard score for example stories, which increase the difficulty of judging the quality of evaluation work we receive from AMT.

Also, as pointed out in \citet{karpinska-etal-2021-perils}, the quality of work from annotators on AMT platform can be of high variance and have poor calibration, therefore, we would obtain more reliable human evaluation results if we hire expert raters such as professional authors or English language teachers.

\section{Inter-annotator Agreement for MTurk Workers}
\label{appdenix:iaa-mturk}

\begin{table}[t]
\centering
\small
\begin{tabular}{p{1ex}cccccc}
\toprule
 & \textbf{IAA} & \textbf{Flu.} &  \textbf{Coh.} & \textbf{Rel.} & \textbf{Log.} & \textbf{Int.}  \\
\midrule
\multirow{2}{3em}{\rotatebox{90}{\textbf{ROC}}} & $r$ & 0.64 & 0.81 & 0.79 & 0.80 & 0.68 \\
& TA & 17.24 & 24.98 & 25.57 & 27.37 & 22.03 \\
 \midrule
 \multirow{2}{3em}{\rotatebox{90}{\textbf{WP}}} & $r$ & 0.51 & 0.70 & 0.74 & 0.71 & 0.54 \\
& TA & 18.37 & 17.01 & 32.65 & 19.73 & 12.93 \\
 \midrule
\multirow{2}{3em}{\rotatebox{90}{\textbf{CNN}}} & $r$ & 0.46 & 0.54 & 0.61 & 0.59 & 0.50 \\
 & TA & 15.13 & 12.61 & 15.97 & 11.76 & 14.29 \\
\bottomrule
\end{tabular}
\caption{Inter Annotator Agreement (IAA) results for each aspect:
fluency (Flu.), coherence (Coh.), relatedness (Rel.), logicality (Log.) and
interestingness (Int.). 
We use one-vs-rest Pearson's $r$ to assess the extent to which each annotator agrees with the consensus. 
Total Agreement (TA) means the percentage where all 3 annotators choose the same score.
}
\label{table:iaa}
\end{table}

We follow \citet{lau-etal-2020-furiously} to estimate one-vs-rest agreement using Pearson's $r$.
For each story,
we single out an annotator's score and 
compare it to the mean scores given by the other two annotators, and we repeat this process for every score in a story and for all stories to compute Pearson's $r$ over the two sets of scores (singled-out scores vs.\ mean scores).
We also compute the percentage where all 3 annotators choose the same score, noting that this is a much stricter agreement metric (as it does not capture the ordinal scale of the scores). Random scoring would produce 4\% for this metric.

IAA results are presented in \autoref{table:iaa}. 
In terms of one-vs-rest agreement ($r$), we find overall good agreement with 9 strong agreement results ($r$ >= 0.6) and 
6 moderate agreement results (0.45 <= $r$ <= 0.6). 
We see some correlation between story length and agreement, as
ROC has the highest agreement (shortest with 5 sentences) and CNN has the lowest (over 20 sentences). 
When it comes to aspects, coherence, relatedness and logicality have higher agreement compared to fluency and interestingness.
While it is intuitive to see interestingness being subjective, fluency is somewhat a surprise.
Manual inspection reveals that annotators have very different standards when it comes to fluency, 
with some workers being more strict about grammar, which contributes to the low agreement.
For total agreement (TA), the numbers range between 10--25\%, which is encouraging as it shows that there is still a good proportion of cases where all annotators agree on a score.



\section{Inter-annotator Agreement for In-house Workers}
\label{appendix:IAA-inhouse}

\begin{table}[t]
\centering
\small
\begin{tabular}{p{1ex}cccccc}
\toprule
 & \textbf{IAA} & \textbf{Flu.} &  \textbf{Coh.} & \textbf{Rel.} & \textbf{Log.} & \textbf{Int.}  \\
\midrule
\multirow{2}{3em}{\rotatebox{90}{\textbf{ROC}}} & $r$ & 0.42 & 0.54 & 0.66 & 0.59 & 0.32 \\
& TA & 38.57 & 25.0 & 25.71 & 25.71 & 8.57 \\
 \midrule
 \multirow{2}{3em}{\rotatebox{90}{\textbf{WP}}} & $r$ & 0.36 & 0.57 & 0.73 & 0.49 & 0.54 \\
& TA & 10.0 & 10.0 & 18.57 & 10.0 & 10.0 \\
 \midrule
\multirow{2}{3em}{\rotatebox{90}{\textbf{CNN}}} & $r$ & 0.36 & 0.41 & 0.47 & 0.37 & 0.35 \\
 & TA & 17.14 & 10.71 & 14.29 & 10.0 & 4.29 \\
\bottomrule
\end{tabular}
\caption{In-house IAA Results.}
\label{table:iaa-inhouse}
\end{table}

The In-house annotation agreement results are shown in \autoref{table:iaa-inhouse}.
Surprisingly, we see similar trends in compared to 
\autoref{table:iaa},
which means our findings are consistent even with a different group of workers on
a different set of stories.

We again see fluency and interestingness are two aspects that have lower agreements.
However, it is interesting that we find fluency has highest TA but rather lower $r$.
Manual inspection reveals that annotators mainly select 4 or 5 for fluency (also can
be seen from high fluency scores for ROC in \autoref{table:he-inhouse}), 
which benefits TA because they have higher chance to agree on the same values, but might
harm $r$ because the score can be negatively influenced by other rare values.
Relevance again has the highest agreement, 
which might be because annotators agree that many stories generated by SOTA models are not related to the given condition. However, compared to IAA results from crowdsource annotators, in-house annotators have lower agreement in general.
we find 2 strong agreement results ($r$ >= 0.6) and 
6 moderate agreement results (0.45 <= $r$ <= 0.6). 



\section{GPT-3 World knowledge Strengths}
\autoref{table:demostories2} and \ref{table:demostories3} demonstrates that GPT-3 can generate better stories than SOTA models thanks to its wide world knowledge strengths.

\begin{table*}[t]
\centering
\begin{tabular}{p{1.7cm} p{13.5cm}} 
\toprule
\textbf{Model} & \textbf{Story}  \\
\midrule
GPT-3 & as soon as i realized that i had died , the first thing that came to mind was how pointless my life had been . i had devoted myself to a cause that was ultimately fruitless and destructive , and in the end i had caused the deaths of millions of people .but then i realized that there was still a chance for redemption . in the after life , i was given the opportunity to travel to different moments in history and try to make things right . i could see the consequences of my actions , and i could try to make different choices that would lead to a better outcome .i started with the holocaust , and i was able to save many people from being killed . but it was difficult , and i had to make a lot of tough choices . eventually i moved on to other moments in history , trying to make things better wherever i could .it was a difficult journey , but it was worth it . i was finally able to make up for the mistakes that i had made in life . \\
\midrule
 PROGEN3 & i knew i was in a bad mood . the blood dripped from my nose and down my chin like the burly balding drifter . was it dark yet ? i could feel my brain becoming small dots of the rotten material . the blood stopped flowing . almost instantly , the world dissolved in my hands . that 's when i noticed something . i was still standing in the middle of the street . i turned to my left and saw the madman .\\ 
\midrule
 HINT & `` hi , `` i said , opening my eyes to a bright light . `` this is hitler . `` i looked at the clock . 5:00 am . `` what 's your name ? `` `` hitler . `` `` hitler . `` `` hitler . `` `` hitler . `` `` hitler . `` \\ 
\midrule
BART & i 'm struck with a fierce sense of shame at that thought . unfortunately it 's been a while since i 've felt this way , and normally it 's * bad * to feel that way , but i think that too should be expected when you 're still alive . there has been so much turmoil since the war . we 've lost children , we 've lost society , we 've even lost ourselves . but now , we 're free . we can finally go on with our lives , like we always should . but something happens that seems wrong , and it 's really hard to forget . there are my reasons for feeling so sad . why will i be mad today ? \\
\bottomrule
\end{tabular}
\caption{Stories generated from GPT-3, HINT, BART and PROGEN3 given the same title ``you 're hitler , and you 've just died . there is an after life . tell me about it ."}
\label{table:demostories2}
\end{table*}

\begin{table*}[t]
\centering
\begin{tabular}{p{1.7cm} p{13.5cm}} 
\toprule
\textbf{Model} & \textbf{Story}  \\
\midrule
GPT-3 & the young man had lost everything . his family , his home , his country . he had nothing left to lose . when he was offered the chance to join isis , he saw it as a way to finally have a purpose in life . he was given a sense of belonging and a sense of worth . he was told that he was fighting for a good cause and that he was helping to protect his people . he was given a sense of identity and a sense of purpose . he was told that he was part of something bigger than himself . \\
\midrule
 PROGEN3 & they 've crippled the ranks of these countries . their leaders not going to the president 's party , everyone having their noses to the grindstone , the war . political parties shooting each other with a firing squad , only to end up murdering each other . `` the one who executed your followers was the first one to hell houng taqibs . `` he 's right , ha is not 'd understand * reason * if he knew that . `` you 're killing my people . ``\\ 
\midrule
 HINT & my eyes stare at the ceiling . it 's just a window , it 's just a window . i 'm not sure if i should close it or close it again . i 'm afraid i ca n't move it . i do n't want to be here . i do n't want to be here . i want to be able to talk to the people who will help me . to be able to hear the voices of people who will help me . to be able to see the lights . to be able to see the people who will help me . \\ 
\midrule
BART & missdamie looked at her young daughter sat with c and o . she 'd put on a radiant glow and bound herself up in an artificial bubble . she was beautiful . she twitched at the vibrancy in her hands . her hair was beautiful and silver . her smile was off . her black nails placed on gentle curves . when she smiled her smile widened and the small eyes of her daughter 's eyes danced . `` isis ? `` i asked , shakily , behind the mask . ms. de los angeles nodded and smiled . `` yes , my name is isis de los angeles , and `` her mother held her hand . `` my son is wondering if isis can cause such a thing . `` \\
\bottomrule
\end{tabular}
\caption{Stories generated from GPT-3, HINT, BART and PROGEN3 given the same title ``make me sympathise with an isis member ."}
\label{table:demostories3}
\end{table*}

\section{Mechanic Turk Instructions}
\begin{figure*}[t]
    \centering
    \includegraphics[width=\textwidth]{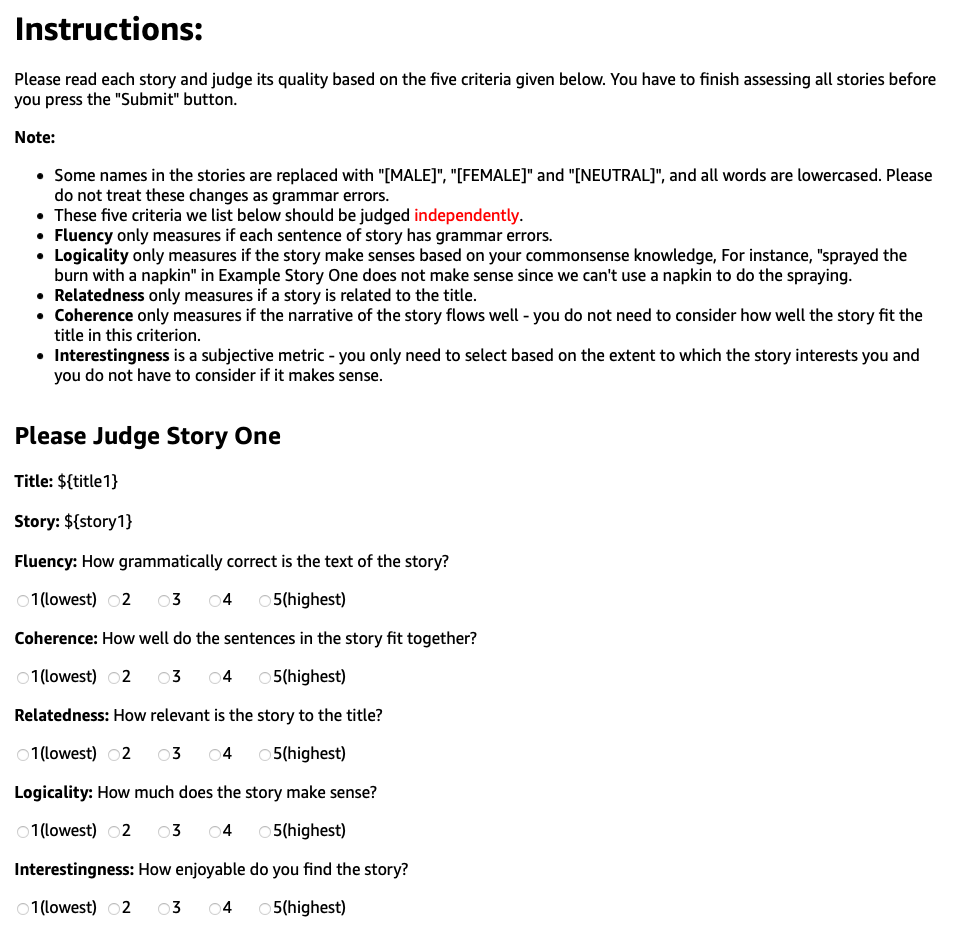}
    \caption{A screenshot of our evaluation questions.}
    \label{fig:AMT}
\end{figure*}

\begin{figure*}[t]
    \centering
    \includegraphics[width=\textwidth]{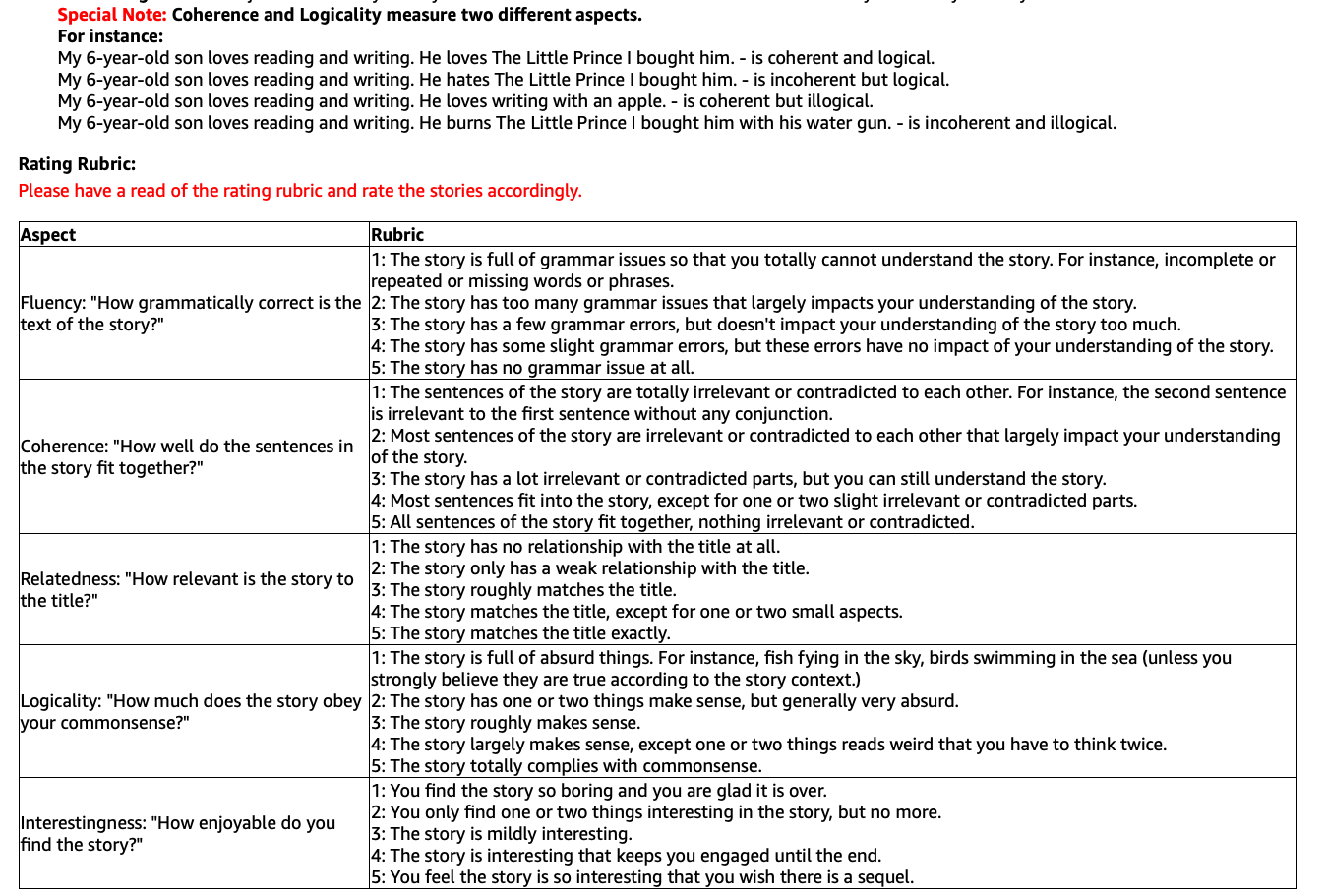}
    \caption{Explanation and Rubric in the Instruction.}
    \label{fig:rubrics}
\end{figure*}

\autoref{fig:AMT} and \ref{fig:rubrics} demonstrate our instructions and rubrics.

\section{Example Prompts}
\autoref{table:promptbasedlearning} shows some examples we use in our prompts for generation.

\begin{table*}[t]
\centering
\begin{tabular}{p{0.05\textwidth} p{0.8\textwidth}} 
 \toprule
  & \textbf{Example} \\ 
 \midrule
 ROC &
 \textbf{prompt}: [MALE] wants to move to a new house .
 \textbf{completion}: his wife [FEMALE] does not want to move . [MALE] convinces her to look at a couple new houses . she does not see any that she likes . [FEMALE] refuses to move .
 \\
 \midrule
 WP &
 \textbf{prompt}: the real reason we 're not supposed to look at the sun during a total eclipse is because it reveals something horrifying .
 \textbf{completion}: the harvest was here . father xander made his way through priory halls toward the sanctuary in the back garden .
 at his advanced years , the trip was a struggle ; each step brought a hefty dose of pain and unpleasant memories .
 it was eclipse day , and while he always prayed for a good outcome , in his eighty years as the priory ’ s eyes , there never was . “ 
the initiates are assembled in the sanctuary , father , ” spoke a fellow priest in the hall . “ thank you , my son , ” father xander replied .
 he paused and looked at the young man . memories came flooding back like sour wine . his name was thomas hampstead , and at thirty years of age ,
 he should have been a robust healthy lad ; but he was not . thomas ’ s skin had a greyish tint and his body was frail .
 \\
 \midrule
 CNN &
 \textbf{prompt}: (CNN) -- Usain Bolt rounded off the world championships Sunday by claiming his third gold in Moscow as he anchored Jamaica to victory in the men's 4x100m relay.
\textbf{completion}: The fastest man in the world charged clear of United States rival Justin Gatlin as the Jamaican quartet of Nesta Carter, Kemar Bailey-Cole, Nickel Ashmeade and Bolt won in 37.36 seconds.
The U.S finished second in 37.56 seconds with Canada taking the bronze after Britain were disqualified for a faulty handover. The 26-year-old Bolt has now collected eight gold medals at world championships,
equaling the record held by American trio Carl Lewis, Michael Johnson and Allyson Felix, not to mention the small matter of six Olympic titles. 
The relay triumph followed individual successes in the 100 and 200 meters in the Russian capital. \"I'm proud of myself and I'll continue to work to dominate for as long as possible,\" Bolt said, 
having previously expressed his intention to carry on until the 2016 Rio Olympics. Victory was never seriously in doubt once he got the baton safely in hand from Ashmeade,
while Gatlin and the United States third leg runner Rakieem Salaam had problems. Gatlin strayed out of his lane as he struggled to get full control of their baton and was never able to get on terms with Bolt. Earlier,
Jamaica's women underlined their dominance in the sprint events by winning the 4x100m relay gold, anchored by Shelly-Ann Fraser-Pryce, who like Bolt was completing a triple. Their quartet recorded a championship record of 41.29 seconds,
well clear of France, who crossed the line in second place in 42.73 seconds. Defending champions, the United States, were initially back in the bronze medal position after losing time on the second handover between Alexandria Anderson and English Gardner,
but promoted to silver when France were subsequently disqualified for an illegal handover. The British quartet, who were initially fourth, were promoted to the bronze which eluded their men's team. Fraser-Pryce,
like Bolt aged 26, became the first woman to achieve three golds in the 100-200 and the relay. In other final action on the last day of the championships, France's Teddy Tamgho became the third man to leap over 18m in the triple jump,
exceeding the mark by four centimeters to take gold. Germany's Christina Obergfoll finally took gold at global level in the women's javelin after five previous silvers, while Kenya's Asbel Kiprop easily won a tactical men's 1500m final.
\\
 \bottomrule
\end{tabular}
\caption{Prompts and GPT-3 Completions, including one example each for ROC, WP and CNN.}
\label{table:promptbasedlearning}
\end{table*}

\section{GPT-3 Stories vs Human References}
\label{sec:GPT-3 Stories vs Human References}
\autoref{table:GPT3-human1}, \ref{table:GPT3-human2} and \ref{table:GPT3-human3} show some examples where GPT-3 obtains higher scores than its human reference.

\begin{table*}[t]
\centering
\begin{tabular}{ p{0.54\textwidth}p{0.07\textwidth}p{0.04\textwidth}p{0.04\textwidth}p{0.04\textwidth}p{0.04\textwidth}p{0.04\textwidth}} 
 \toprule
 \textbf{Title [SEP] Story} & \textbf{From} & \textbf{Flu.} & 
 \textbf{Coh.} & \textbf{Rel.} & \textbf{Log.} & \textbf{Int.} \\
 \midrule
tell me the story of a totally inconsequential character who does n’t realize he ‘s wrapped up in a massive high-stakes plotline - and keep them in the dark about it . [SEP] ivan was just an ordinary guy who led an ordinary life . he worked a dead-end job at a local factory , and spent his evenings and weekends playing his favorite online game with his friends . one day , out of the blue , ivan was contacted by a representative of the game company , who offered him a chance to participate in a closed beta test for their new game . ivan was excited to participate and eagerly agreed to the terms and conditions .little did he know that he was about to become embroiled in a global conspiracy . the game company was actually a front for a secret organization that was planning to use the game to recruit new members . ivan was the perfect candidate , because he was someone who was looking for an escape from his ordinary life .the organization was planning to use the game to recruit new members from all over the world , and they were counting on ivan to help them . ivan had no idea what he was getting himself into , and the organization was careful to keep him in the dark about their true intentions . ivan thought he was just playing a game , but he was actually a pawn in a much larger scheme . & GPT-3 & 5.0 & 5.0 & 5.0 & 5.0 & 5.0 \\
 \midrule
tell me the story of a totally inconsequential character who does n’t realize he ‘s wrapped up in a massive high-stakes plotline - and keep them in the dark about it . [SEP] someone was moving around on screen three . * probably just another junkie stumbling around . fucking dumbass . * barry did n’t really give a shit what the junkie was up to anyway . his job did n’t pay enough for that . it ’s not like they thought he was actually all that important , anyway . a security guard ? minimum wage , take your paycheck , shut up , and move along , thank you very much . you have to sit for twelve hours in the dead of night and devote all of your attention to watching the same twelve tiny boxes ? that ’s nice . how ’s \$ 7.25 an hour sound ? you ’re preventing vandalism and break-ins ? how wonderful . would you like a cookie ? & human & 3.67 & 2.33 & 2.0 & 3.33 & 2.33 \\

 \bottomrule
\end{tabular}
\caption{GPT-3 Generated Story Obtains Higher Scores than Its Human Reference.}
\label{table:GPT3-human1}
\end{table*}

\begin{table*}[t]
\centering
\begin{tabular}{ p{0.54\textwidth}p{0.07\textwidth}p{0.04\textwidth}p{0.04\textwidth}p{0.04\textwidth}p{0.04\textwidth}p{0.04\textwidth}} 
 \toprule
 \textbf{Title [SEP] Story} & \textbf{From} & \textbf{Flu.} & 
 \textbf{Coh.} & \textbf{Rel.} & \textbf{Log.} & \textbf{Int.} \\
 \midrule
you meet the grim reaper at a crossroads . you are fearful as he approaches , but notice he seems just as fearful of you . [SEP] the grim reaper was a fearful figure , draped in a cloak of blackness with a face that was hidden in shadow . as he came closer , you could see the fear in his eyes . `` why are you so afraid of me ? `` you asked . `` because you 're the first person to ever see me like this , `` he replied . `` i 'm not here to take you , i 'm here to ask for your help . `` & GPT-3 & 4.0 & 4.67 & 4.67 & 3.67 & 3.67 \\
 \midrule
you meet the grim reaper at a crossroads . you are fearful as he approaches , but notice he seems just as fearful of you . [SEP] my eyes jump from snowbank to snowbank , one featureless as the next . the road i walk upon is a dark grey slush , the result of many cars and snow plows making their way through the void . yet i had not seen a car in miles . i shiver . but not from the cold . the compelling . it had awakened me . this ominous feeling that i had to be at a certain place at a certain time . it controlled me . it made me get dressed . it made me open the door . it had even made me ignore my wife , who i usually kiss goodbye . the road seemed endless , but i felt no reason to stop . & human & 3.33 & 3.33 & 1.67 & 2.33 & 2.33 \\
 \bottomrule
\end{tabular}
\caption{GPT-3 Generated Story Obtains Higher Scores than Its Human Reference.}
\label{table:GPT3-human2}
\end{table*}

\begin{table*}[t]
\centering
\begin{tabular}{ p{0.54\textwidth}p{0.07\textwidth}p{0.04\textwidth}p{0.04\textwidth}p{0.04\textwidth}p{0.04\textwidth}p{0.04\textwidth}} 
 \toprule
 \textbf{Title [SEP] Story} & \textbf{From} & \textbf{Flu.} & 
 \textbf{Coh.} & \textbf{Rel.} & \textbf{Log.} & \textbf{Int.} \\
 \midrule
you are able to see the number of days a person has left to live above their heads . for the first time in your life , you see someone with a negative number . [SEP] as you walk the streets , dodging the rushing crowds , you notice something different about one of the people walking by . their number is displayed in a negative value . for a moment , you feel a pang of sadness and empathy . you know what that means - that person is going to die soon . you watch as they go about their day , completely unaware of the timer ticking away above their head . you wonder who they are , what they 're doing , and who will be there with them when their time runs out . & GPT-3 & 4.67 & 4.33 & 4.67 & 4.0 & 5.0 \\
 \midrule
you are able to see the number of days a person has left to live above their heads . for the first time in your life , you see someone with a negative number . [SEP] crowds are the worst . frank squints through the flurry of hovering digits as he shoulders his way through the masses of shoppers . the cacophony of vociferous patrons , distressed employees , and typical black friday commotion crescendos around him . he pauses to watch two shoppers vie over the last remaining television monitor ( only \$ 600 ! ) . one is a blonde woman whose pigtails have been jostled into disarray ; above her hovers the number 10951 . the other is n't as fortunate ; the number 582 glows above her greying hair . frank sighs . he 's sure there 's something … vaguely poignant about fighting over a product priced at a greater value than your remaining days of life . socio-rhetorical commentary on modern consumerism , perhaps . he has little time to contemplate , however , as another shopper impatiently shoves him forward , and frank resumes his pace . while he walks , frank studies the various faces passing by : a woman in an elegant , blue hijab ; a portly , bearded man ; a fatigued mother clutching her two children by the hands ... . & human & 3.67 & 3.0 & 3.33 & 4.0 & 3.33 \\
 \bottomrule
\end{tabular}
\caption{GPT-3 Generated Story Obtain Higher Scores than Its Human Reference.}
\label{table:GPT3-human3}
\end{table*}

\section{Example Generations}
\autoref{table:randomgeneration} shows randomly picked generation example for each dataset.

\begin{table*}[t]
\centering
\begin{tabular}{p{0.2\textwidth} p{0.65\textwidth}} 
 \toprule
 \textbf{Title} & \textbf{Generation}  \\
 \midrule
the boy cried . & the boy cried because he was sad and scared . his parents were getting a divorce and he did n't know what was going to happen to him . he was going to have to leave his home and his friends and move to a new house . his mother said that he could choose where he wanted to live , but his father said that he had to live with him . the boy did n't know what to do .\\
 \midrule
 your extreme apathy has been mistaken as spiritual enlightenment , and now people are building a religion around you . & 
my name is john and i have a problem . well , actually , a lot of people have a problem with me but that 's not the point . the point is that i don 't care about anything . at all . 
not the way most people care about things anyway . when most people see a homeless person on the street they feel compassion and want to help . 
when i see a homeless person on the street i think `` well , there goes another lazy bum who 's too good for work `` . and i 'm not joking when i say that most people see the world in a completely different way than i do . they see the good in people and want to help them . 
i see the bad in people and want to help them . see the difference ? i 'm not saying that my way of seeing the world is wrong and theirs is right . i 'm just saying that we see the world differently and that 's okay . 
or at least it should be . but it 's not because people can 't accept that i don 't care about anything . 
they think that i 'm enlightened because i don 't feel the pain and suffering that they do . 
they think that because i don 't care about anything i must be in a better place than they are .and that 's why they 're building a religion around me . \\
 \midrule
(CNN) -- Hurricane Ida moved into the southern Gulf of Mexico Sunday, prompting a declaration of emergency in Louisiana and concern along the U.S. Gulf Coast.
& ida Louisiana Gov. Bobby Jindal declared a state of emergency Saturday as the Category 1 hurricane churned in the Gulf of Mexico, according to a news release from the governor's office. 
"Based on the latest forecasts, Hurricane Ida is expected to make landfall in Louisiana as a Category 1 hurricane and we must be prepared for significant impacts," Jindal said in the release. 
"I have directed my team to do everything possible to protect the lives and property of the people of Louisiana." 
The hurricane could bring storm surges, high winds and heavy rains to the coast, the National Hurricane Center said. 
The storm was about 220 miles (355 kilometers) south-southeast of the mouth of the Mississippi River on Sunday morning and 
was moving north-northwest at about 9 mph (15 kph), the center said. The hurricane had maximum sustained winds of 75 mph (120 kph). 
Forecasters said Ida could strengthen a bit before making landfall. The center of the storm is expected to come ashore early Tuesday morning. \\
 \bottomrule
\end{tabular}
\caption{Randomly Chosen GPT-3 Generation Examples from Each Dataset.}
\label{table:randomgeneration}
\end{table*}

\section{Soft Plagiarism Examples}
\label{sec:Soft Plagiarism Examples}
\autoref{table:similaritycheck} shows some GPT-3 generated news stories where we find true online stories.

\begin{table*}[t]
\centering
\begin{tabular}{p{0.85\textwidth}} 
 \toprule
     \textbf{GPT-3 Generated Story (Link to the Online Resource) } \\
 \midrule
Washington (CNN) -- With the posting of 400,000 classified documents from the Iraq war, WikiLeaks has shown a much heavier hand redacting compared to its previous publication of documents. In an interview with Wolf Blitzer on Wednesday, WikiLeaks founder Julian Assange said that the organization redacted the names of Iraqis who provided information to the United States military in order to protect them from retaliation."We have a duty to protect some of the people who are named in these documents," Assange said.WikiLeaks has redacted the names of coalition forces and informants, as well as the names of Iraqis who cooperated with the United States military, in all of its recent releases of classified documents.  (\url{https://www.reuters.com/article/britain-assange-idUSL4N2GD2PZ)}\\
 \midrule
 (CNN) -- A former University of Virginia lacrosse player convicted of killing his ex-girlfriend was sentenced to 23 years in prison on Thursday, according to the Commonwealth Attorney's office in Charlottesville, Virginia. George Huguely V was convicted in February of second-degree murder in the May 2010 death of Yeardley Love. He was also convicted of grand larceny and burglary.Prosecutors had asked for a sentence of 30 years, while Huguely's defense attorneys had asked for 10 years."Yeardley was my everything," Huguely said in a statement before sentencing. "I was not thinking, I was not rationale, and I was not in control of my own actions."Love was found beaten to death in her off-campus apartment. Huguely's DNA was found on Love's body and in her bedroom.During the trial, prosecutors said Huguely had sent Love a series of emails in which he threatened to kill her."I'm going to kill you. You deserve it. You took the one thing I loved the most," Huguely wrote in one email, according to prosecutors.
 (\url{https://en.wikipedia.org/wiki/Murder_of_Yeardley_Love})
 \\
 \midrule
(CNN) -- Crisis club Racing Santander has been thrown out of next season's Copa del Rey after its players staged a bizarre on-pitch boycott of a quarterfinal tie against Real Sociedad. Racing Santander's players staged a bizarre on-pitch boycott of a Copa del Rey quarterfinal tie against Real Sociedad on Wednesday, leading to their expulsion from the competition.The players refused to take to the field in the first leg of the last-eight tie at the Anoeta Stadium in San Sebastian, protesting at the club's failure to pay their wages.After a delay of around 45 minutes, the referee finally called off the game, with the Spanish Football Federation (RFEF) confirming Racing Santander's expulsion from the competition."The Competition Committee has decided to expel Racing Santander from the Copa del Rey for not appearing in the match," said an RFEF statement."Real Sociedad have been awarded a 3-0 win and will now face the winners of the match between Real Madrid and Valencia, who will play the semi-final."Racing Santander, who are currently bottom of the Spanish first division, released a statement on their website protesting at the club's treatment by the RFEF."Racing Santander has today been expelled from the Copa del Rey for not appearing in the match against Real Sociedad, despite the fact that our players were prepared to play," said the statement."We believe that this decision is unjust, given that our players have not been paid their wages for several months and have made it clear that they would not be able to compete in the match."The club also released a video of the players refusing to take to the field, with captain David Lafuente leading the protest.Racing Santander's expulsion from the Copa del Rey means they will not compete in next season's competition, while Real Sociedad will face the winners of the match between Real Madrid and Valencia in the semi-final.
(\url{http://edition.cnn.com/2014/01/30/sport/football/spanish-cup-boycott-santander-football/index.html})
\\
 \bottomrule
\end{tabular}
\caption{GPT-3 generated CNN News and link to the Online Resource}
\label{table:similaritycheck}
\end{table*}

\section{GPT-3 Generation Issues}
\autoref{table:badgenerations} shows some GPT-3 generated stories with several issues.

\begin{table*}[t]
\centering
\begin{tabular}{p{2cm} p{3cm} p{9cm}} 
 \toprule
 \textbf{Issue} & \textbf{Title} & \textbf{Generation} \\ [0.5ex] 
 \midrule
 Generate Chinese from an English title  & you are an alien body snatcher , and this is your first day in your new host body . you have no idea what his life is . & 你不是人，你是外星人。这是你第一天在新的主人身上。你不知道他的生活是什么。 你只是偶然得到他的身体，他的生活已经开始了，而你还不知道这是什么。你的主人是一个普通人，他没有你的外星人能力。你不知道你的主人在做什么，也不知道他有没有你的朋友。 你只是在他的身体里，完全不知道他的生活。\\
 \midrule
 Generate Japanese from an English title & what it means to live ( 250 words ) . & 
 人生とは 何かを探して生きることである。探し続けることで、生きる価値を見つける。 例えば夢を探し続けることで、その夢が叶う確率を高める。夢を叶えるために努力をし続けることで、その夢が叶う可能性が高まる。 しかし、夢を叶えるために努力をし続けることは難しい。夢を叶えるためには、その夢に対して情熱を持つことが大切だ。 そして、その夢を叶えるためには、自分を信じることが大切だ。夢を叶えるためには、その夢にチャレンジすることが大切だ。そして、夢を叶えるためには、失敗を恐れずに前進することが大切だ。 \\
 \midrule
 Stick word together to create ``new" words & write a short story based off the lyrics to any song . & there was a timewhen we were youngand our dreams were alivewe wanted to be rock starsand travel the worldwe wanted to be someonesomeone famoussomeone lovednow we 're just oldour dreams have diedour passions gonewe just want to be left aloneto die in peace\\
  \midrule
 Copy from the leading text & [MALE] was a very sweet dog . & [MALE] was a very sweet dog . he loved his family and was always happy to see them . even though he was old , he never stopped wagging his tail and licking everyone 's faces . \\
  \midrule
  Copy from the prompt & [MALE] is a manager at kfc . & he started as a fry cook [MALE] wants to move to a new house , but his wife [FEMALE] does not want to move . [MALE] convinces her to look at a couple new houses , and she does not see any that she likes . [FEMALE] refuses to move . (\autoref{table:promptbasedlearning}) \\
\midrule
Generate profanity words & [FEMALE] was eating lunch at school . & a boy came up to her and asked her to go out with him . she said no and he called her a stuck up b * * * * .\\
 \bottomrule
\end{tabular}
\caption{Examples of some interesting generation errors we find for GPT-3.}
\label{table:badgenerations}
\end{table*}

\section{Pearson Correlations between Each Story Aspects}
\begin{figure*}[t]
     \centering
     \begin{subfigure}[b]{0.6\textwidth}
         \centering
         \includegraphics[width=\textwidth]{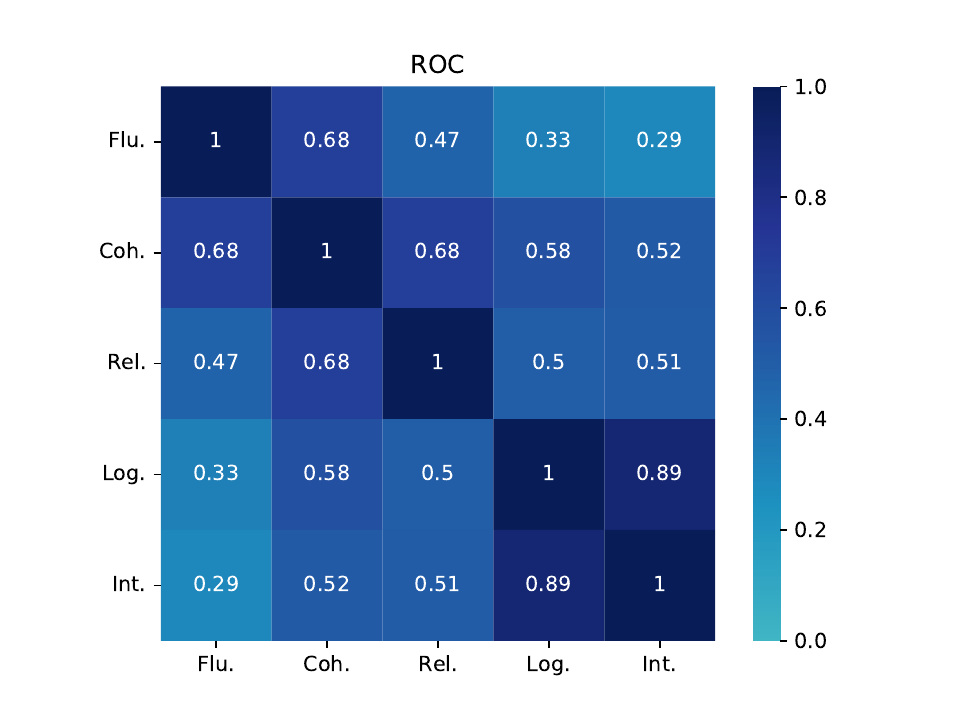}
         \caption{ROC}
         \label{fig:roc}
     \end{subfigure}
     \hfill
     \begin{subfigure}[b]{0.6\textwidth}
         \centering
         \includegraphics[width=\textwidth]{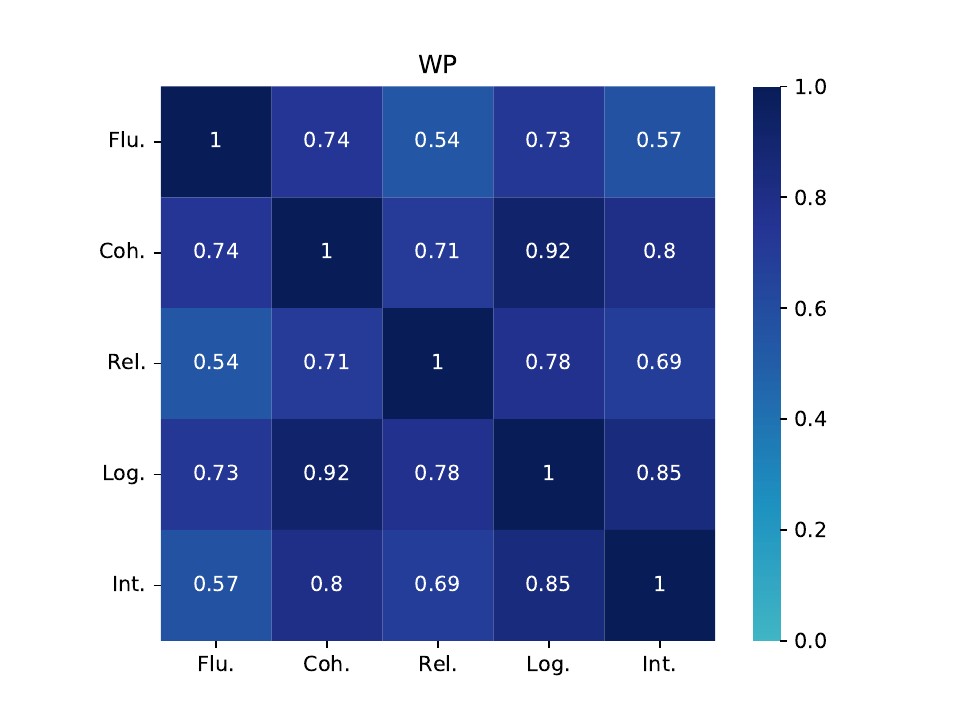}
         \caption{WP}
         \label{fig:wp}
     \end{subfigure}
     \hfill
     \begin{subfigure}[b]{0.6\textwidth}
         \centering
         \includegraphics[width=\textwidth]{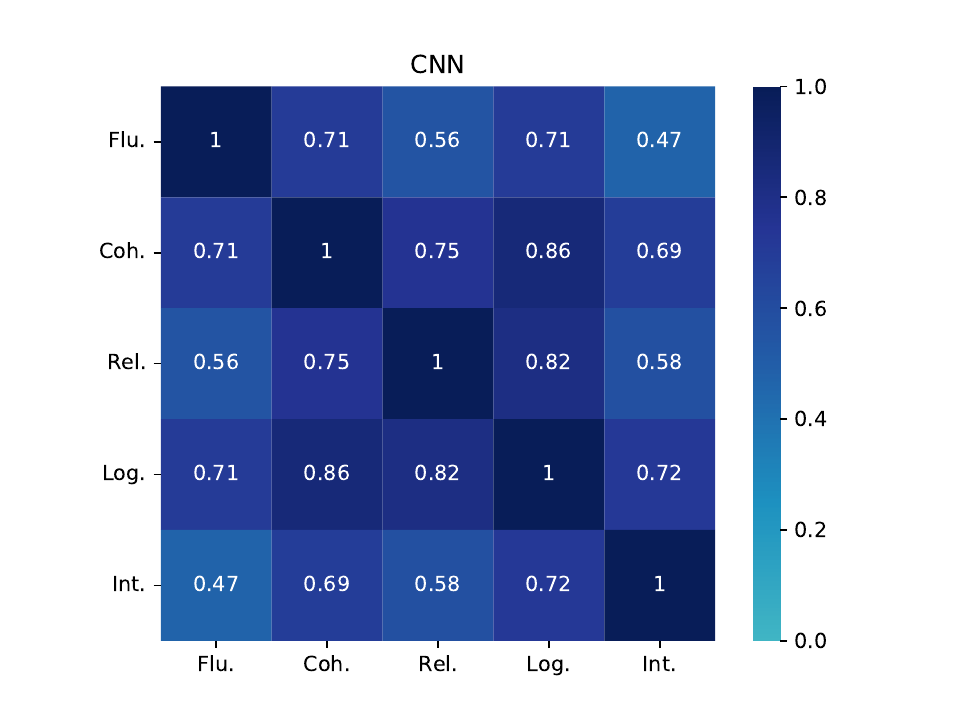}
         \caption{CNN}
         \label{fig:cnn}
     \end{subfigure}
        \caption{Pearson Correlations between Each Aspect from Crowdsourcing annotations.}
        \label{fig:correlations-turk}
\end{figure*}

\begin{figure*}[t]
     \centering
     \begin{subfigure}[b]{0.6\textwidth}
         \centering
         \includegraphics[width=\textwidth]{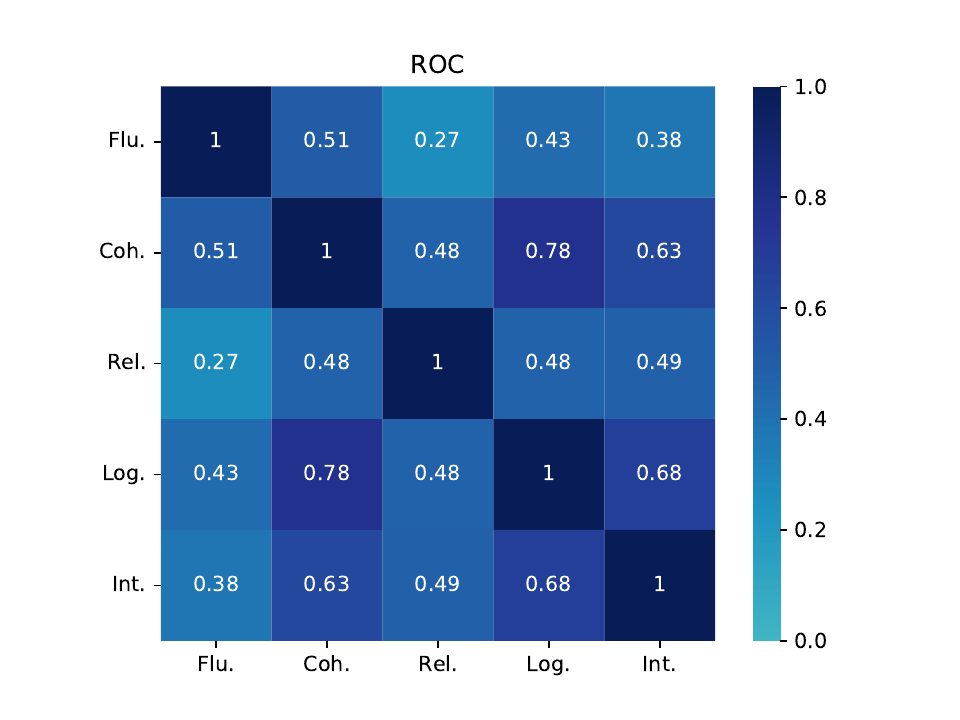}
         \caption{ROC}
         \label{fig:roc-in-house}
     \end{subfigure}
     \hfill
     \begin{subfigure}[b]{0.6\textwidth}
         \centering
         \includegraphics[width=\textwidth]{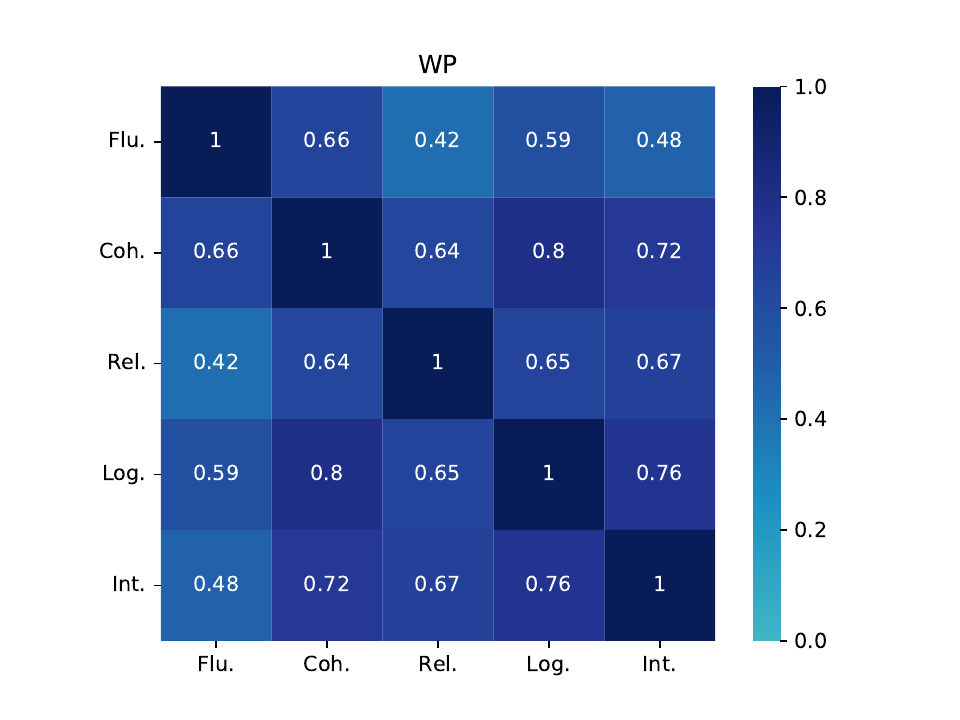}
         \caption{WP}
         \label{fig:wp-in-house}
     \end{subfigure}
     \hfill
     \begin{subfigure}[b]{0.6\textwidth}
         \centering
         \includegraphics[width=\textwidth]{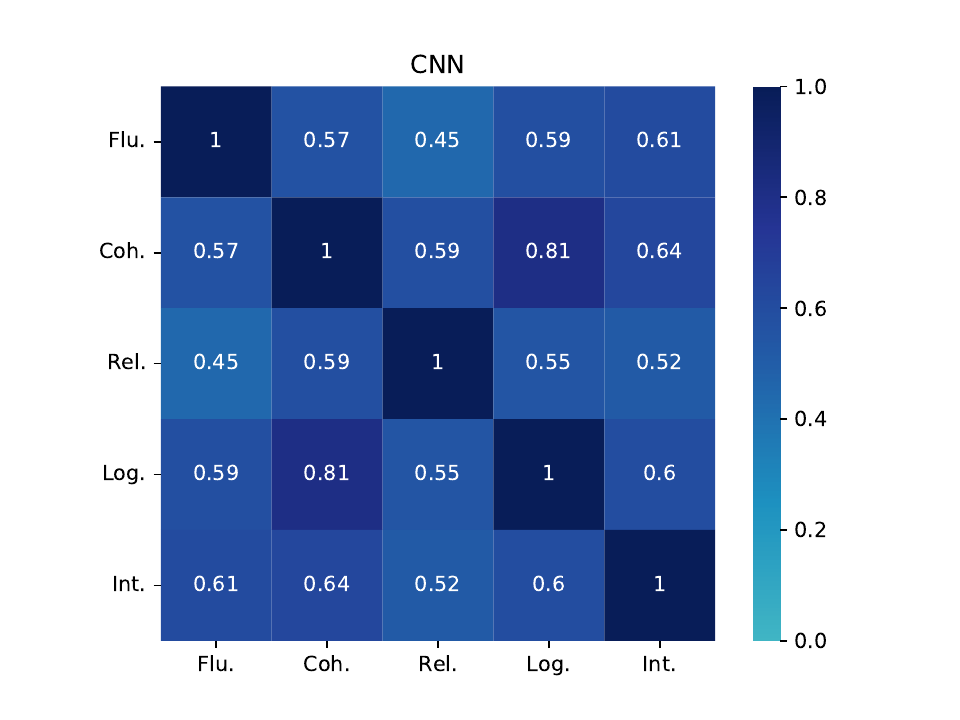}
         \caption{CNN}
         \label{fig:cnn-in-house}
     \end{subfigure}
        \caption{Pearson Correlations between Each Aspect from in-house annotations.}
        \label{fig:correlations}
\end{figure*}

\autoref{fig:correlations-turk} and \ref{fig:correlations} present Pearson Correlations between Each Story Aspects for MTurk and in-house workers respectively.

\end{CJK*}
\end{document}